\pdfoutput=1

\documentclass[11pt]{article}

\newif\ifarxiv
\arxivtrue 

\ifarxiv
    \usepackage{acl}
\else
    \usepackage[review]{acl}
\fi

\usepackage{times}
\usepackage{latexsym}
\usepackage{amsmath}
\usepackage{amssymb}
\usepackage{stmaryrd}

\usepackage[T1]{fontenc}

\usepackage[utf8]{inputenc}

\usepackage{microtype}

\usepackage{inconsolata}

\usepackage{graphicx}

\usepackage{enumitem}
\usepackage{array}
\usepackage{amsthm}
\usepackage{booktabs} 

\usepackage{listings}
\usepackage{longtable}
\usepackage{booktabs}
\usepackage{soul,xcolor}
\usepackage{longtable}
\sethlcolor{yellow}
\usepackage{subcaption}
\usepackage{multirow}
\usepackage{bm}
\usepackage{svg}

\usepackage{calligra}
\newcommand{\specialfont}[1]{{\itshape\calligra #1}}
\usepackage{tikz}
\newcommand{\grayb}{\textcolor{gray}{50.00}}
\newcommand{\grayz}{\textcolor{gray}{0.00}}
\definecolor{taborange}{rgb}{1.0, 0.498, 0.0549}
\definecolor{tabblue}{rgb}{0.1216, 0.4667, 0.7059}
\newcommand{\orangedotline}{%
    \tikz[baseline=-0.5ex]{
        \draw[thick, taborange] (0,0) -- (0.6,0);
        \filldraw[orange] (0.3,0) circle (0.07cm);
    }
}

\newcommand{\orangedottedline}{%
    \tikz[baseline=-0.5ex]{
        \draw[thick, taborange, dashed] (0,0) -- (0.6,0);
        \filldraw[orange] (0.3,0) circle (0.07cm);
    }
}

\newcommand{\bluexline}{%
    \tikz[baseline=-0.5ex]{
        \draw[thick, tabblue] (0,0) -- (0.6,0);
        \draw[thick, tabblue] (0.2,-0.08) -- (0.4,0.08);
        \draw[thick, tabblue] (0.4,-0.08) -- (0.2,0.08);
    }
}

\newcommand{\bluedottedxline}{%
    \tikz[baseline=-0.5ex]{
        \draw[thick, tabblue, dashed] (0,0) -- (0.6,0);
        \draw[thick, tabblue] (0.2,-0.08) -- (0.4,0.08);
        \draw[thick, tabblue] (0.4,-0.08) -- (0.2,0.08);
    }
}

 \usepackage[most]{tcolorbox}

\newtcolorbox{myquotebox}{
  colback=white!0, 
  colframe=black, 
  rounded corners,
  boxrule=0.5pt, 
  title=Prompt:,
  left=2mm, 
  right=2mm, 
  top=1mm, 
  bottom=1mm 
}

\usepackage{todonotes}
\usepackage{xspace}
\newcommand{\fixme}[2][]{\todo[color=goldenrod,size=\scriptsize,fancyline,caption={},#1]{#2}} 
\newcommand{\note}[4][]{\todo[author=#2,color=#3,size=\scriptsize,fancyline,caption={},#1]{#4}} 
\newcommand{\mrinmaya}[2][]{\note[#1]{mrinmaya}{green}{#2}\xspace}
\newcommand{\jingwei}[2][]{\note[#1]{jingwei}{yellow}{#2}\xspace}

\definecolor{lightgrey}{RGB}{158, 158, 158}
\definecolor{goldenrod}{rgb}{0,0,0.8}
\definecolor{deepred}{rgb}{0.6,0,0}
\definecolor{deepgreen}{rgb}{0,0.5,0}
\definecolor{pink}{RGB}{219, 48, 122}
\definecolor{forestgreen}{RGB}{34,139,34}
\definecolor{goldenrod}{RGB}{218,165,32}
\definecolor{sepia}{RGB}{112,66,20}

\usepackage[capitalise]{cleveref}
\crefname{figure}{Fig.}{Figs.}
\crefname{table}{Table}{Tables}
\crefname{appendix}{App.}{App.}
\crefname{section}{§}{§§}
\crefname{equation}{Eq.}{Eqs.}

\newcommand{\algname}{UNIT}

\newcommand{\smallgreen}[1]{\textcolor{deepgreen}{\scriptsize #1}}
\newcommand{\smallred}[1]{\textcolor{red}{\scriptsize #1}}
\definecolor{lightred}{RGB}{254, 138, 138}
\definecolor{lightblue}{RGB}{176, 195, 248}
\definecolor{lightgreen}{RGB}{138, 218, 174}
\newcommand\myparagraph[1]{
\vskip 0.05in 
\noindent{\bf {#1}}}
\newcommand*\samethanks[1][\value{footnote}]{\footnotemark[#1]}

\definecolor{boxborder}{RGB}{86, 113, 209}  
\definecolor{boxbg}{RGB}{255, 255, 255}    
\definecolor{boxtitle}{RGB}{255, 255, 255} 
\definecolor{boxheader}{RGB}{86, 113, 209}  

\makeatletter
\g@addto@macro\normalsize{%
  \setlength{\abovedisplayskip}{4pt}
  \setlength{\belowdisplayskip}{4pt}
  \setlength{\abovedisplayshortskip}{4pt}
  \setlength{\belowdisplayshortskip}{4pt}
}
\makeatother

\usepackage{tikz} 
\usepackage{xcolor} 

\newcommand{\bcircle}[1]{%
    \tikz[baseline=(char.base)]{
        \node[shape=circle,fill=black,draw,inner sep=1.5pt] (char) 
        {\textcolor{white}{\textbf{#1}}};}}

\title{AI for Climate Finance: Agentic Retrieval and Multi-Step Reasoning for Early Warning System Investments}

\author{ 
    \textbf{Saeid Ario Vaghefi}\textsuperscript{\rm 1,2}\thanks{Equal Contributions.},
    \textbf{Aymane Hachcham}\textsuperscript{\rm 1}\samethanks\\
    \textbf{Veronica Grasso}\textsuperscript{\rm 2},
    \textbf{Jiska Manicus}\textsuperscript{\rm 2}\\
    \textbf{Nakiete Msemo}\textsuperscript{\rm 2},
    \textbf{Chiara Colesanti Senni}\textsuperscript{\rm 1}\\
    \textbf{Markus Leippold}\textsuperscript{\rm 1,3} \\
    \textsuperscript{\rm 1}University of Zurich \hspace{5mm}
    \textsuperscript{\rm 2}WMO \hspace{5mm}
    \textsuperscript{\rm 3}Swiss Finance Institute (SFI) \\
    \texttt{\{saeid.vaghefi, aymane.hachcham, chiara.colesantisenni, markus.leippold\}@df.uzh.ch} \\
    \texttt{\{svaghefi, vgrasso, jmanicus, nmsemo\}@wmo.int}
}

\begin{document}
\maketitle
\begin{abstract}
Tracking financial investments in climate adaptation is a complex and expertise-intensive task, particularly for Early Warning Systems (EWS), which lack standardized financial reporting across multilateral development banks (MDBs) and funds which are the main funders of these EWS projects. Analysts regularly encounter diverse PDF files containing tables and images with inconsistent formatting, rows, and columns, making it difficult and time-consuming to analyze reports and extract proper financial information. 
To address this challenge, we introduce an agent-based Retrieval-Augmented Generation (RAG) system that orchestrates contextual retrieval with internal chain-of-thought (COT) reasoning to extract relevant financial data, classify investments, and ensure compliance with funding guidelines. 
Our study focuses on a real-world application: tracking EWS investments funded by the Climate Risk and Early Warning Systems (CREWS) Fund. 
We evaluate our agent-based RAG pipeline on 25 MDB project documents from the CREWS Fund, 
comparing it against five model candidates—(1) a Zero-Shot Classifier (Baseline), (2) a Few-Shot “Zero Rule” Classifier, (3) a fine-tuned transformer-based classifier, and (4) a Few-Shot-V2 CoT+ICL classifier—across both multi-label classification and budget allocation tasks. 
Our agent-based RAG achieves 87\% accuracy, 89\% precision, and 83\% recall, significantly outperforming these benchmarks. 
We also benchmark it against the Gemini 2.0 Flash AI Assistant, setting the stage for a comparative study of Glass-Box Agents versus Black-Box Assistants to quantify the benefits of an agentic pipeline in transparency, explainability, and performance. 
Finally, we release a benchmark dataset and expert-annotated corpus to catalyze further research in AI-driven climate finance tracking.\footnote{We will open-source all code, LLM generations, and human annotations. This can foster further innovation and development in this important area, leading to even more sophisticated and effective tools for managing climate finance.}
\end{abstract}

\section{Introduction} \label{sec:introduction}
Recent advances in Large Language Models (LLMs) have transformed investment tracking, financial reporting, and compliance monitoring in climate finance. However, tracking financial flows and categorizing investments in Early Warning Systems (EWS) remains challenging due to the lack of standardized structures and terminologies in financial reports from Multilateral Development Banks (MDBs) and climate funds.  

\myparagraph{Motivation.}  
Early Warning Systems (EWS) are essential for disaster risk reduction and climate resilience. 
The United Nations (UN) has prioritized universal EWS access by 2027 through its Early Warnings for All (EW4All) initiative, 
emphasizing that timely warnings reduce economic losses and save lives. 
Studies show that 24 hours of advance warning can reduce damages by 30\%, while every dollar invested in early warning systems saves up to ten 
dollars in avoided losses\footnote{See Appendix \ref{app:EWS} for more on EWS.}. Despite their importance, 
EWS investments lack financial transparency, as MDB reports often fail to classify and track funding allocations systematically. The lack of standardized financial reporting for EWS investments by MDBs and funds creates inefficiencies and hinders effective resource allocation. 

In this work, we frame investment tracking as a multi-label classification task—each text or table snippet may belong to one or more of the CREWS Fund’s pillars—and, once labels are assigned, we automatically extract budget allocations with grounding evidence spans directly from the PDF. The resulting output is a structured JSON mapping each pillar to its supporting evidence and allocated funds, vastly reducing the time and expertise required for manual review.
To make our task concrete, we adopt the following pillar definitions:
\begin{itemize}
    \item \textbf{Pillar 1, Disaster risk knowledge:} Comprehensive information on hazards, exposure, vulnerability, and capacity—including the production, rescue, sharing, and application of risk data to inform early action.
    \item \textbf{Pillar 2, Hazard detection and forecasting:} Non-structural capacity‐building and structural infrastructure for multi-hazard monitoring, analysis, forecasting, and data management (e.g., observing networks, forecasting models, radars).
    \item \textbf{Pillar 3, Warning dissemination and communication:} Non-structural systems and structural platforms (cell-broadcast, sirens, SMS, social media, TV/radio, public address) that ensure timely, people-centered delivery of warnings to all at-risk groups.
    \item \textbf{Pillar 4, Preparedness to respond:} Non-structural planning and training (contingency, anticipatory action, public education) alongside structural shelters and resource centers that translate warnings into life-saving measures.
    \item \textbf{Cross-Pillar, Governance and sustainability:} Cross‐cutting institutional arrangements, policy frameworks, stakeholder coordination, and financial planning necessary to sustain and scale the four core pillars.
\end{itemize}
\myparagraph{Context.}  
EW4All underscores the need for financial transparency in climate adaptation: clear tracking of fund flows can improve project monitoring and reduce disaster losses. Proper monitoring also makes it possible to identify where investments have been made compared to other areas, which pillars have received funding, and which aspects have been under-invested. This insight enables better resource allocation and ensures that all critical components of climate adaptation are adequately supported. However, MDB financial reports present a highly heterogeneous mix of structured tables, free-form text, and institution-specific jargon, without standardized categorization or terminology.
Classical NLP approaches-e.g. fine-tuned transformer classifiers or rule-based table parsers-are brittle in this setting, requiring extensive labeled data to cover every layout variation and often failing to generalize across documents \cite{karpukhin-etal-2020-dense}, \cite{chen2020tabfactlargescaledatasettablebased}.
Even layout-aware transformers (LayoutLM \cite{10.1145/3394486.3403172}, Longformer \cite{beltagy2020longformerlongdocumenttransformer}) assume some consistency  in formatting or demand expensive layout annotations.

\begin{figure*}[htp]
\centering	\includegraphics[width = 1\textwidth]{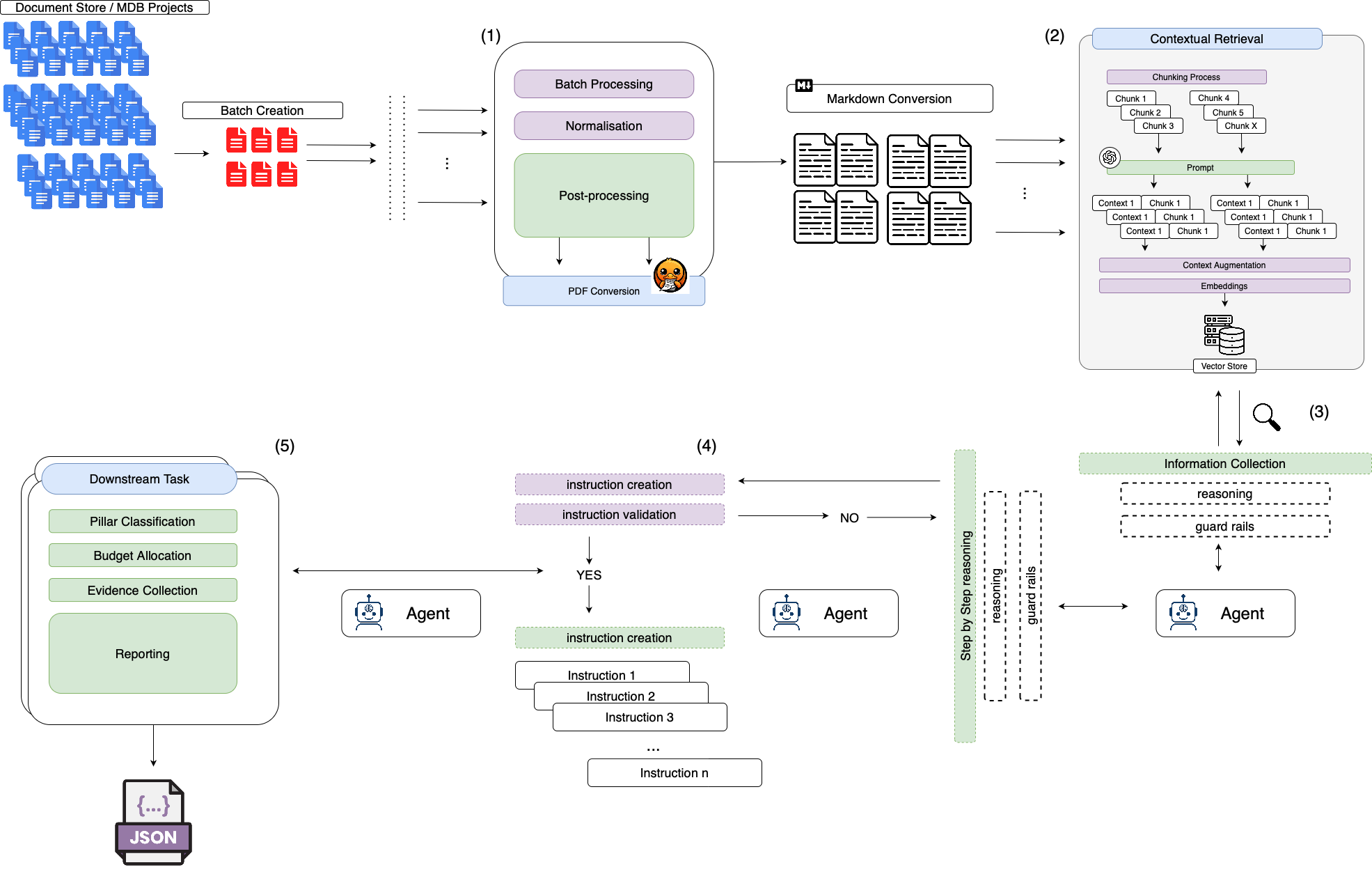}
\vspace{-0.3em}
\caption{AI-driven financial tracking pipeline for EWS investments. The different steps are: (1) PDF conversion, (2) context retrieval, (3) information storage and collection, (4) iterative sub-query and instruction creation, (5) dowstream task execution (pillar classification and budget allocation).}
\label{fig:pipeline}
\vspace{-0.7em}
\end{figure*}

To address these challenges, we argue that a multi-stage AI information system is essential.
By decomposing the task into dedicated components (c.f. Section~\ref{sec:methodology}, Figure~\ref{fig:pipeline}),
the pipeline can robustly handle diverse reporting formats, minimize annotation needs, and produce fully grounded, compact JSON outputs. 
This modular design leverages the strengths of each subcomponent to deliver the most reliable and scalable solution for climate finance transparency.

\myparagraph{Contribution.}  
We introduce the EW4All Financial Tracking AI-Assistant, an agent-based RAG pipeline that employs multi-modal extraction—parsing text, tables, and graphs—and internal chain-of-thought reasoning with built-in guardrails to produce robust, explainable decision chains across multiple sub-tasks.
We benchmark this approach against 4 model candidates—Zero-Shot Classifier (Baseline), Few-Shot “Zero Rule” Classifier, Fine-Tuned Transformer Classifier, and a Few-Shot-V2 CoT+ICL Classifier—on 25 CREWS-Fund documents, where it achieves 87\% accuracy, 89\% precision, and 83\% recall, 
a 23\% lift over traditional NLP methods.
We extend our evaluation to include the Gemini 2.0 Flash AI Assistant, setting up the first systematic contrast between transparent, agentic pipelines (Glass-Box Agents) and end-to-end black-box systems—quantifying gains in transparency, expert validation, and classification performance.
Finally, we open-source our expert-annotated corpus, benchmark dataset, and all prompt designs to catalyze future AI-driven climate finance tracking research.

\paragraph{Implications.}  
By improving climate finance transparency, this AI-driven approach provides structured, evidence-based insights into MDB investments. 
The integration of retrieval-augmented generation and agentic AI enhances decision-making, financial accountability, and policy formulation in global climate investment tracking. 
With a clearer understanding of investment patterns, gaps, and overlaps, stakeholders can make more informed decisions regarding resource allocation, project prioritization, and policy formulation in global climate investment tracking. 
The integration of retrieval-augmented generation (RAG) and agentic AI also enhances explainability and expert validation, making the system's outputs more reliable for decision-making. 
The evidence-based insights provided by the AI system can support the formulation of more effective climate adaptation policies. By identifying areas where investments are lacking or where funding guidelines might need adjustments, policymakers can use this information to optimize resource allocation for climate resilience. Hence, this work contributes to broader AI applications in climate finance, supporting international initiatives that seek to optimize resource allocation for climate resilience.

\section{Related Literature}





RAG improves knowledge-intensive tasks by integrating external retrieval with LLM generation \cite{lewis2020retrieval}, yet traditional RAG remains limited by static retrieval pipelines. Agentic RAG enhances adaptability by incorporating iterative retrieval and decision-making, improving factual accuracy and multi-step reasoning \cite{xi2023risepotentiallargelanguage, yao2023react, guo2024largelanguagemodelbased}. Multi-agent frameworks extend this by refining retrieval for applications such as code generation and verification \cite{guo2024largelanguagemodelbased, liu2024largelanguagemodelbasedagents}, advancing explainability and human-AI collaboration.  

In-Context Learning (ICL) allows LLMs to generalize from few-shot demonstrations without fine-tuning \cite{brown2020language}, but its effectiveness hinges on example selection. Retrieval-based ICL improves prompt efficiency, and reward models further refine in-context retrieval \cite{wang2024reinforcement}. CoT prompting facilitates step-by-step reasoning, significantly boosting performance in arithmetic and commonsense tasks \cite{wei2022chain, kojima2022large}. Self-consistency decoding enhances CoT by aggregating multiple reasoning paths \cite{wang2023selfconsistencyimproveschainthought}, while example-based prompting strengthens complex question-answering capabilities \cite{diao2024activepromptingchainofthoughtlarge}.


\section{Methodology}
\label{sec:methodology}
MDB project documents are characterized by highly heterogeneous layouts—mixed narrative text, nested tables, multi-column formats, footnotes, and embedded figures-such
that evidence of EWS pillars and funding may be dispersed across pages, tables, and descriptive passages.
Conventional retrieval or single-pass parsing pipelines struggle to (i) locate semantically related spans when they reside in separate structural regions,
(ii) reconcile duplicate or overlapping budget figures across distinct table formats and (iii) ensure end-to-end consistency in the face of OCR errors or layout ambiguities.

To address these challenges, we adopt an agent-based retrieval-augmented generation (RAG) framework that orchestrates:
\begin{enumerate}
    \item \textit{Iterative sub-query generation}, where an LLM-driven agent dynamically decomposes the overall extraction task into fine-grained retrieval instructions.
    \item \textit{Hybrid semantic-lexical search}, combining dense vector retrieval with BM25-style keyword matching to capture both contextual relevance and exact matches.
    \item \textit{Self-validation loops or guardrails}, in which the agent examines the sufficiency and coherence of retrieved chunks (re-issuing queries when coverage thresholds are unmet).
    \item \textit{Schema-aware consolidation}, formatting the final evidence spans and associated numeric allocations into a single structured JSON output.
\end{enumerate}

Figure~\ref{fig:pipeline} illustrates the overall pipeline with all its components.

\subsection{Embedding Construction and Indexing}

Effective downstream reasoning over MDB PDFs requires a robust embedding index that reconciles heterogeneous layouts and scattered evidence. 
To this end, we employ a unified four-stage pipeline that breaks the task into four main components: document parsing, chunking, context augmentation, embedding generation, and vector storage.
First, we extract both raw text and structural elements from each document \(d\) using the Docling document converter (Auer et al., 2024):
\begin{equation}
T_d = \text{DoclingParser}(d),
\label{eq:doclingparse}
\end{equation}
where \(T_d\) denotes the set of all extracted elements (text, tables, images) from document \(d\).
We then partition \(T_d\) into three disjoint chunk sets,
\begin{equation}
\mathcal{C} = \mathcal{C}_{\mathrm{table}} \cup \mathcal{C}_{\mathrm{text}} \cup \mathcal{C}_{\mathrm{image}},
\label{eq:chunksets}
\end{equation}
where $\mathcal{C}_{\mathrm{table}}$, $\mathcal{C}_{\mathrm{text}}$, and $\mathcal{C}_{\mathrm{image}}$ denote the sets of table, text, and image chunks, respectively. Where \(\mathcal{C}_{\mathrm{table}}\) comprises automatically detected table regions, \(\mathcal{C}_{\mathrm{text}}\) contains narrative passages split at markdown‐style headers, and \(\mathcal{C}_{\mathrm{image}}\) includes embedded figures. Writing \(\mathcal{C}=\mathcal{C}_{\mathrm{table}}\cup\mathcal{C}_{\mathrm{text}}\cup\mathcal{C}_{\mathrm{image}}\), this decomposition prevents loss of context and mitigates parsing errors arising from complex multi‐column layouts and mixed content.

Next, to situate each chunk within its document context and reduce semantic ambiguity \cite{günther2024latechunkingcontextualchunk}, 
we generate a concise, two-sentence summary for each \(c\in\mathcal{C}\). We prompt an LLM with \(P_{\mathrm{ctx}}(c,T_{d})\) to obtain
\begin{equation}
\mathrm{ctx}(c)\;=\;\mathrm{LLM}\bigl(P_{\mathrm{ctx}}(c,T_{d})\bigr),
\label{eq:context}
\end{equation}
and form the augmented chunk
\begin{equation}
c' = c \;\oplus\; \mathrm{ctx}(c).
\label{eq:augchunk}
\end{equation}
By anchoring each chunk to its global narrative, we ensure that subsequent retrieval captures both fine‐grained detail and overall document significance.

We encode each augmented chunk \(c'\) into two modality‐specific latent spaces: one jointly for text and tables, and one for images. Formally, we define  
\begin{equation}
e_{\mathrm{tt}}(c') = f_{\mathrm{tt}}(c') \in \mathbb{R}^{d_{\mathrm{tt}}},\quad
e_{\mathrm{im}}(c') = f_{\mathrm{im}}(c') \in \mathbb{R}^{d_{\mathrm{im}}}
\label{eq:embeddings}
\end{equation}

where \(f_{t}\) is a joint text-table encoder trained to capture both narrative and structured tabular semantics, and \(f_{im}\) is an image encoder \cite{radford2021learningtransferablevisualmodels} 
specialized for visual feature extraction. 
We index these two embedding spaces in Weaviate environment by defining separate NamedVector configurations—one for text-table properties and another for image properties—thus 
preserving modality-specific semantics and enabling efficient hybrid (semantic + lexical) search across modalities. 
At query time, Weaviate dispatches each multimodal query to the appropriate vector index and returns the top-$k$ relevant chunks for downstream RAG orchestration.

This embedding step condenses high-dimensional text and layout features into a semantic space where related content remains in proximity.

Finally, each embedding \(e(c')\) is stored in a vector database with metadata \(\mathrm{meta}(c') = \{\texttt{file\_name}:f\}\), where \(f\) is the PDF’s filename:
\begin{equation}
\mathrm{VDB\_store}\bigl(e(c'),\,\mathrm{meta}(c')\bigr).
\label{eq:vdbstore}
\end{equation}

At inference time, for a given file ID \(f\) and query \(q\), we retrieve the top-5 semantically and lexically relevant chunks via
\begin{equation}
\mathcal{R}(f) \;=\; \mathrm{VDB\_query}(q,\,f), 
\quad |\mathcal{R}(f)|=5
\label{eq:vdbquery}
\end{equation}

\section{Hybrid Retrieval via Rank Fusion}

In addition to the above procedure, we employ a hybrid search strategy that combines dense vector search with BM25F-based keyword search \cite{robertson2009probabilistic} to leverage both semantic similarity and exact lexical matching. Let $\mathcal{R}_v(q, f)$
denote the set of candidate chunks retrieved via dense vector search, and let $\mathcal{R}_k(q, f)$ denote the candidate chunks obtained via BM25F keyword search. To fuse these two retrieval sets, we use Reciprocal Rank Fusion (RRF) \cite{cormack2009reciprocal}. For each candidate chunk \( c \in \mathcal{R}_v(q, f) \cup \mathcal{R}_k(q, f) \), we compute an RRF score as:
\begin{equation}
\text{RRF}(c) = \sum_{i\in\{v,k\}} \frac{1}{\text{rank}_i(c) + K},
\label{eq:rrf}
\end{equation}
where \( \text{rank}_i(c) \) is the rank of \( c \) in retrieval system \( i \) (with lower ranks corresponding to higher relevance) and \( K \) is a smoothing constant (typically set to 60). The final set of retrieved chunks is then given by selecting the top five candidates according to their RRF scores:
\begin{equation}
\mathcal{R}(f) = \operatorname{Top5}\Bigl(\mathcal{R}_v(q, f) \cup \mathcal{R}_k(q, f), \, \text{RRF}(c)\Bigr).
\label{eq:hybridtop5}
\end{equation}
This hybrid method harnesses the semantic sensitivity of dense vector retrieval alongside the precise lexical matching of BM25F, thereby enhancing the overall disambiguation and retrieval performance during downstream processing.

\subsection{Classification and Budget Allocation}

For each retrieved chunk \( c' \in \mathcal{R}(f) \), we apply the following four methods to classify the chunk (i.e., assign it a class \( y \) from the five pillars) and to allocate an associated budget \( B \).

\subsubsection{Zero-Shot and Few-Shot Classification}

In this approach, we construct a prompt \( P_{\text{Class+Budget}}(c') \) that includes the content of the augmented chunk and, in the few-shot setting, several annotated examples. The LLM is then queried to simultaneously produce an outcome classification \( y \) and an associated budget \( B \):
\begin{equation}
\{y, B\} = \mathrm{LLM}(P_{\text{Class+Budget}}(c')).
\label{eq:zsfs}
\end{equation}
This method leverages the pre-trained knowledge of the LLM, with few-shot prompting guiding its responses.

\subsubsection{Fine-Tuned Transformer-Based Classifier}

In another approach, we fine-tune a transformer-based classifier \( M_{\text{ft}} \) on a labeled dataset \( \{(c'_i, y_i)\}_{i=1}^{N} \). The model is used to classify each augmented chunk:
\begin{equation}
y = M_{\text{ft}}(c').
\label{eq:ftclassifier}
\end{equation}
Subsequently, an LLM is used to determine the budget allocation for each class. The prompt \( P_{\text{Budget}}(c', y) \) is constructed using the chunk and its classification:
\begin{equation}
B = \mathrm{LLM}(P_{\text{Budget}}(c', y)).
\label{eq:ftbudget}
\end{equation}
The final result for each chunk is the tuple \(\{y, B\}\).

\subsubsection{Few-Shot-V2: Chain-of-Thought (CoT)}

This approach employs a three-step Chain-of-Thought (CoT) strategy, resulting in a tuple \(\{y, B\}\):
\begin{enumerate}[label=\arabic*., nosep]
    \item \textbf{Reformatting:} If \( c' \) represents a table, it is reformatted into a clean markdown table:
    \begin{equation}
    c'' = \mathrm{LLM}(P_{\text{reformat}}(c')).
    \label{eq:cotreformat}
    \end{equation}
    Otherwise, we set \( c'' = c' \).
    \item \textbf{Classification:} A classification prompt is used to classify the (reformatted) chunk:
    \begin{equation}
    y = \mathrm{LLM}(P_{\text{Class}}(c'')).
    \label{eq:cotclass}
    \end{equation}
    \item \textbf{Budget Allocation:} A subsequent prompt allocates the budget:
    \begin{equation}
    B = \mathrm{LLM}(P_{\text{Budget}}(c'', y)).
    \label{eq:cotbudget}
    \end{equation}
\end{enumerate}

\subsubsection{Agent-Based Approach}

This method uses an agent that follows a sequence of instructions and performs RAG queries:
\begin{enumerate}[label=\arabic*., nosep]
    \item \textbf{Instruction Generation:} The agent, primed with examples of annotated PDFs and the desired output format, generates a list of sub-task instructions \( I = \{i_1, i_2, \ldots, i_k\} \) to complete the classification and budget allocation task. It also generates a list of queries \( Q = \{q_1, q_2, \ldots, q_l\} \) to use if the sub-tasks require querying the vector database.
    \item \textbf{Sub-Task and Query Mapping:} The agent maps instructions \( I\) to queries \( Q\).
    \item \textbf{Sub-Task Execution:} For each instruction \( i_j \), if the sub-task requires querying the vector database, a retrieval is performed to extract relevant chunks:
    \begin{equation}
    c'_{i_j} = \mathrm{VDB\_query}(q_{i_j}, f).
    \label{eq:agentquery}
    \end{equation}
    \item \textbf{Sub-Task Validation:} The agent performs a self-healing step to validate that the retrieved chunks \(c'_{i_j}\) are sufficient. If not, a new query \(q_{i_j}^{\mathrm{new}}\) is generated and the retrieval is repeated:
    \begin{equation}
    {c'}_{i_j}^{\mathrm{final}} =
    \begin{cases}
    \mathrm{VDB\_query}(q_{i_j}^{\mathrm{new}}, f), & \\\text{if } c'_{i_j} \text{ is insufficient}, \\[1ex]
    c'_{i_j}, & \text{otherwise}.
    \end{cases}
    \label{eq:agentselfheal}
    \end{equation}
    \item \textbf{Final Formatting:} After finishing all the sub-tasks, the final step formats the output as JSON:
    \begin{equation}
    \{y, B\} = \mathrm{LLM}(P_{\text{Format}}(\{\mathrm{result}_{I}\})).
    \label{eq:agentfinal}
    \end{equation}
\end{enumerate}


\section{Results}

\subsection{Pillar-Level Budget Classification}

We frame the CREWS-Fund experiment as a joint pillar-classification and budget-allocation task.
For every document $d$ we observe a \emph{budget vector}:
\begin{equation}
  \mathbf{b}_d = \bigl(b_{d,1},\dots, b_{d,5}\bigr) \in \mathbb{R}_{\ge 0}^{5}, \qquad
  \sum_{p=1}^{5} b_{d,p} = B^{\mathrm{tot}}_d,
  \label{eq:budgetvector}
\end{equation}
where $b_{d,p}$ denotes the amount assigned to EWS pillar $p$ and $B^{\mathrm{tot}}_d$ is the document’s total EWS envelope.

We derive binary pillar indicators as
\begin{equation}
  y_{d,p} = \llbracket b_{d,p} > 0 \rrbracket \in \{0,1\},
  \label{eq:indicator}
\end{equation}
where $\llbracket \cdot \rrbracket$ denotes the Iverson bracket, which is 1 if the condition is true and 0 otherwise. Eventually, our model outputs $\hat{\mathbf{b}}_d$ and $\hat{y}_{d,p} = \llbracket \hat{b}_{d,p} > 0 \rrbracket$.

A prediction for pillar $p$ in document $d$ is counted as a \emph{true positive} (TP) only if \textbf{both} conditions hold:
\begin{enumerate}[label=(\alph*)]
  \item \textbf{Correct label.}  
        The model assigns the pillar label that is truly present, i.e., $y_{d,p} = 1$ and $\hat{y}_{d,p} = 1$.\\
        (The task is multi-label over the fixed set of five EWS pillars.)
  \item \textbf{Budget fidelity.}  
        The predicted allocation is numerically faithful, i.e.,
        \begin{equation}
          \bigl|\hat{b}_{d,p} - b_{d,p}\bigr| \le 0.05\,B^{\mathrm{tot}}_d,
          \label{eq:budgetfidelity}
        \end{equation}
        a $\pm5\%$ tolerance around the gold amount for that pillar.
\end{enumerate}

Using \(25\) CREWS-Fund PDFs, we benchmark five baselines
(Zero-Shot, Few-Shot, Transformer, Few-Shot-CoT) against our
\emph{Glass-Box Agentic} pipeline.
Table~\ref{tab:results} reports the scores:
the agent attains \textbf{0.87 accuracy, 0.89 precision, 0.83 recall},
an (8-14) pp lift over the strongest baseline.

\begin{table}[ht]
    \centering
    \resizebox{\columnwidth}{!}{%
    \begin{tabular}{lccc}
    \toprule
    \textbf{Method} & \textbf{Accuracy} & \textbf{Precision} & \textbf{Recall} \\
    \midrule
    Zero-Shot       & 0.41 & 0.40 & 0.61 \\
    Few-Shot        & 0.42 & 0.45 & 0.64 \\
    Transformer     & 0.41 & 0.64 & 0.32 \\
    Few-Shot-CoT    & 0.51 & 0.63 & 0.71 \\
    Agent           & \textbf{0.87} & \textbf{0.89} & \textbf{0.83} \\
    \bottomrule
    \end{tabular}
    }
    \caption{Evaluation metrics for budget distribution across the EWS Pillars.}
    \label{tab:results}
    \end{table}

These figures show that the agent not only identifies the correct set of
pillars but also assigns budget to them with tight numeric fidelity,
providing a solid reference line for the broader Glass-Box vs. Black-Box
study in \cref{sec:glass-vs-black} (see Figure\ref{fig:outupt} for a sample analytic
report)

\subsection{Glass-Box \emph{vs.} Black-Box Study (MDB Evidence Set)}
\label{sec:glass-vs-black}

To test whether transparency still pays off in a truly end-to-end setting,
we build a second benchmark: an \emph{annotated corpus of 500 evidence
segments} extracted from multi-layout MDB project documents,
co-curated with World Meteorological Organization (WMO).
Each segment is labelled with
(i) its EWS pillar,
(ii) the budget amount assigned to that pillar,
(iii) the evidence–pillar linkage, and
(iv) the document’s total EWS budget.
This allows us to probe retrieval quality, reasoning traceability and
numerical fidelity in a single pass.

We compare three systems: \textbf{Glass-Box Agent} (The agentic system explain in section 4.1.4) to \textbf{Gemini 2.0 Flash},
a Black-Box assistant that processes the same PDF via a single prompt, and \textbf{OpenAI Assistants}
another Black-Box baseline likewise queried end-to-end.

\subsection{Prompt Engineering for Gemini 2.0 Flash EWS Financial Analysis}
Our prompt design strategy employs a modular architecture with clearly delineated components. We structure the prompt with five key segments: (1) a role definition establishing the AI as a financial analyst specialized in Early Warning Systems, (2) project-specific goals directing the analysis toward EWS funding allocation, (3) a comprehensive taxonomy reference that standardizes EWS classification, (4) methodical analysis instructions with explicit calculation guidance, and (5) a structured JSON output format ensuring consistency across analyses. This hierarchical decomposition transforms a complex financial assessment task into a sequence of manageable analytical steps, promoting both thoroughness and traceability.

Performance is analysed along five facets; the metrics listed below are
computed for \emph{each} system:

\begin{description}
    \item[Evidence extraction.]
          We measure how well a system retrieves the gold evidence
          segments.  
          Key metrics include
          \emph{Recall} (\(\tfrac{\text{TP}}{\text{TP+FN}}\)),
          \emph{Precision} (\(\tfrac{\text{TP}}{\text{TP+FP}}\)),
          their harmonic mean \(F_{1}\),
          and \(\text{Recall}@5\), the fraction of gold segments found
          within the top-5 ranked results.
  
    \item[Amount distribution across pillars.]
          For every evidence–pillar pair the system predicts an amount
          \(\hat{b}_{d,p}\).  
          Accuracy, Precision, Recall and \(F_{1}\) are computed under a
          \(\pm5\%\) tolerance with respect to the gold amount
          \(b_{d,p}\) (cf.\ Eq.~\eqref{eq:budgetfidelity}). where \( \hat{b}_{d,p} \) is the predicted allocation, \( b_{d,p} \) is the gold allocation, and \( B^{\mathrm{tot}}_d \) is the total budget for document \( d \). The prediction is considered correct if it falls within \(\pm 5\%\) of the true value.

    \item[Pillar-label assignment.]
          The task is multi-label over the five EWS pillars.
          We calculate per-pillar TP, FP and FN, then aggregate macro-
          averages of Accuracy, Precision, Recall and \(F_{1}\).
  
    \item[Evidence-to-label mapping.]
          A mapping is correct if
          (a) the evidence segment is retrieved and
          (b) it is linked to the correct pillar.  
          Metrics follow the same TP/FP/FN template as above.
  
    \item[Total EWS amount prediction.]
          After summing predicted pillar amounts, we compare the total
          \(\hat{B}^{\text{tot}}_d\) against the gold
          \(B^{\text{tot}}_d\) using absolute accuracy and percentage
          error.
  \end{description}

\subsection{Interpretation of the benchmark}
\begin{figure*}[t]
    \centering
    \includegraphics[width=0.55\linewidth]{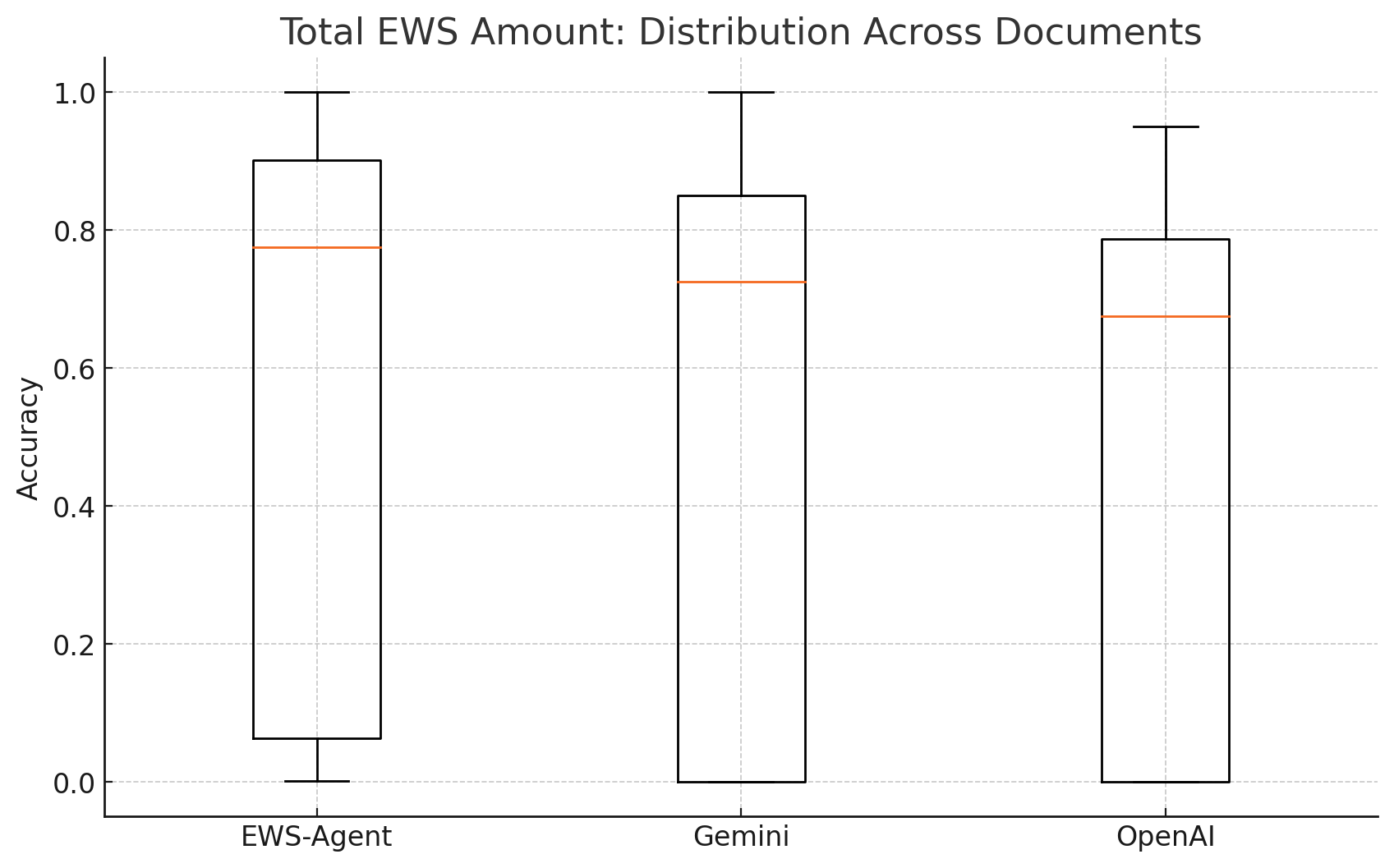}
    \hfill
    \includegraphics[width=0.44\linewidth]{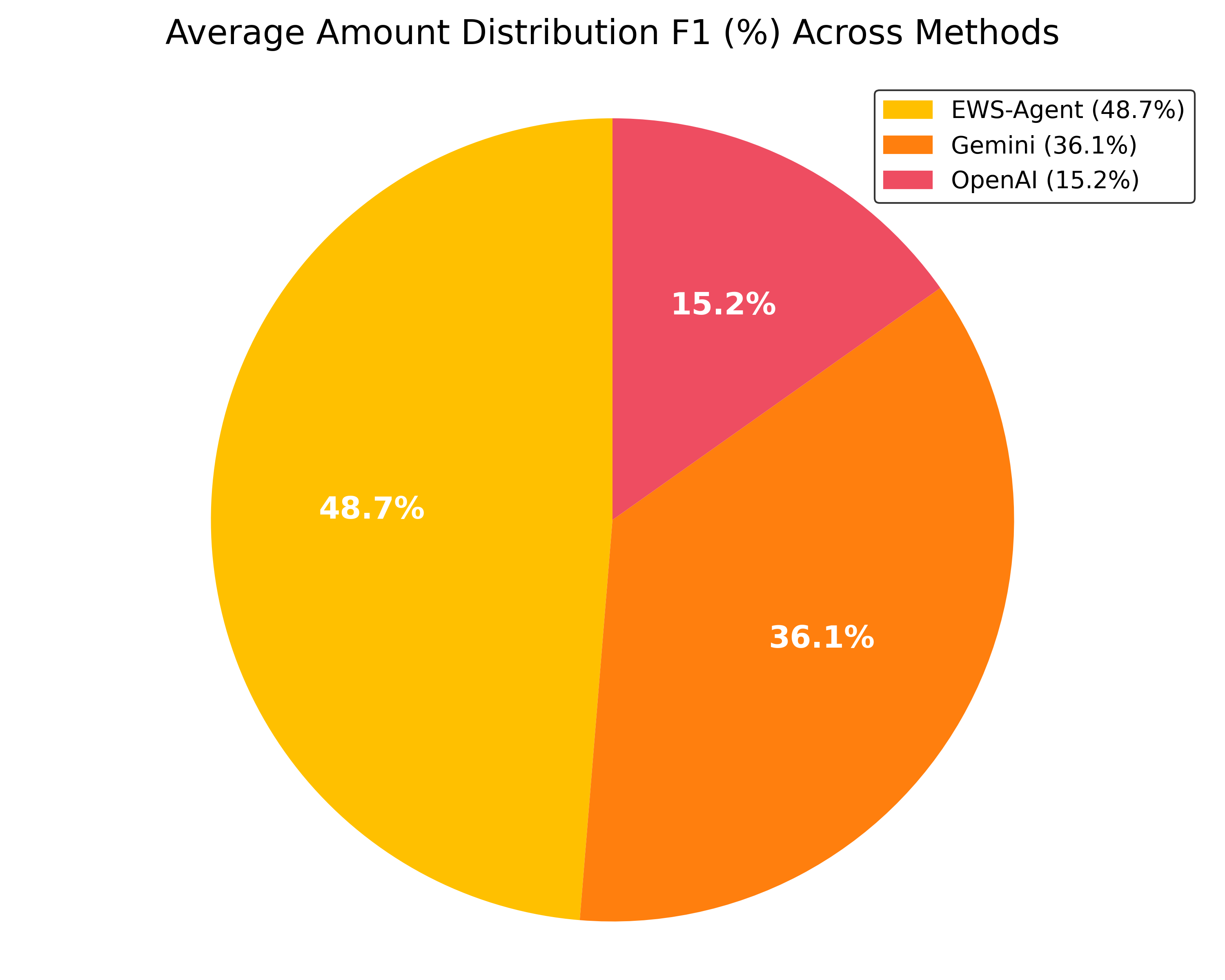}
    \caption{Left: distribution of \emph{total-amount accuracy} for the 500-document
             MDB set.
             Right: share of the \emph{macro-averaged F\textsubscript{1}} obtained by
             each system on the \textbf{amount-per-pillar} task.}
    \label{fig:box-pie}
  \end{figure*}
  
  \begin{figure}[t]
    \centering
    \includegraphics[width=\linewidth]{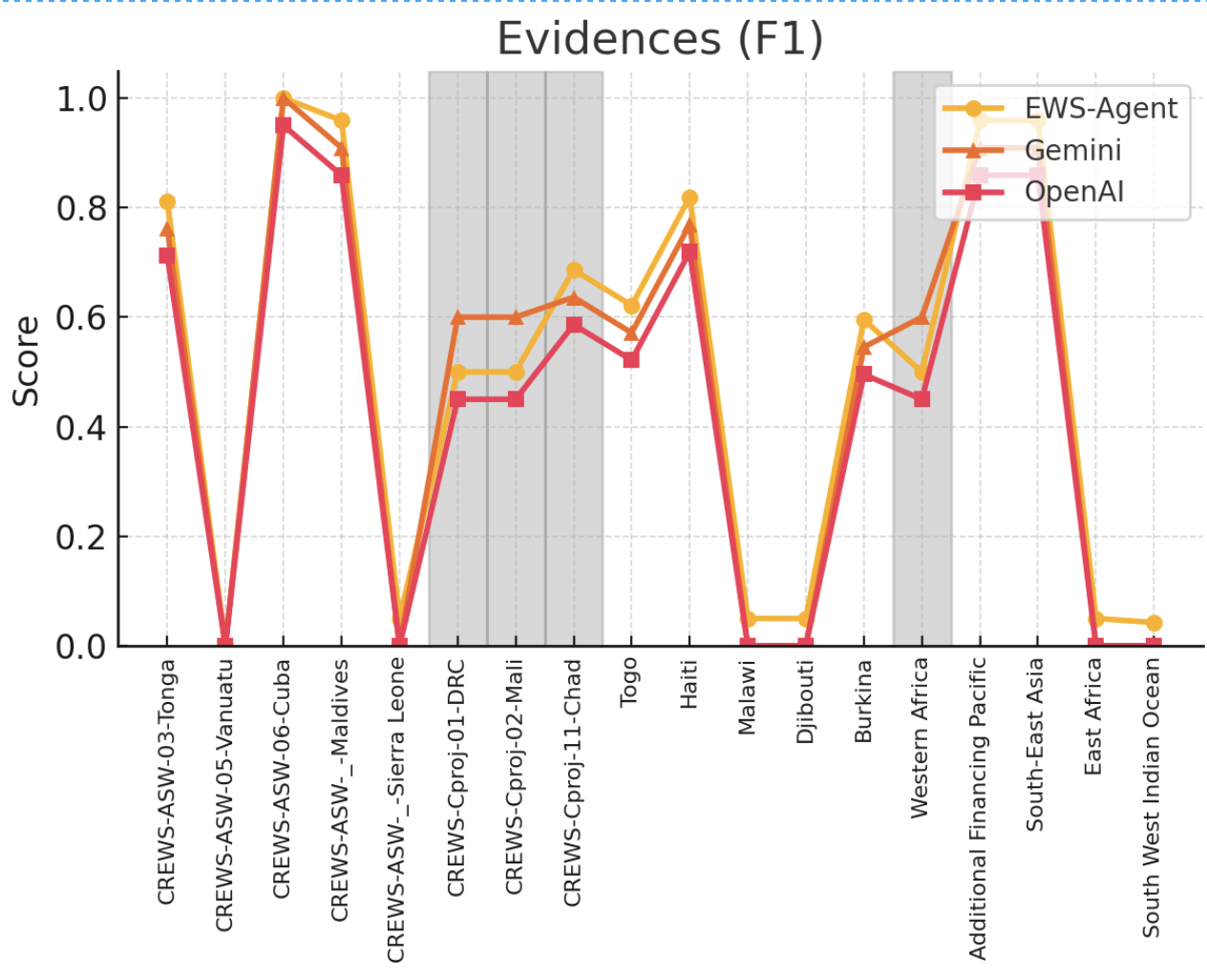}
    \caption{Per-document F\textsubscript{1} for \textbf{evidence extraction}.
             Grey bands highlight projects in which budget figures are dispersed
             across narrative sections rather than formatted tables.}
    \label{fig:scatter-evidence}
  \end{figure}

  \paragraph{Total-amount accuracy (Fig.~\ref{fig:box-pie}, left).}
  The Glass-Box Agent attains the highest median accuracy
  (\(\tilde{x}\approx0.78\)) and a narrow inter-quartile range,
  demonstrating both precision and stability across heterogeneous layouts.
  Gemini and OpenAI trail behind (median \(\approx0.72\) and
  \(\approx0.68\), respectively) and exhibit heavier tails, indicating more
  frequent large errors.
  
  \paragraph{Amount-per-pillar performance (Fig.~\ref{fig:box-pie}, right).}
  When the accuracy metric is tightened to pillar-level allocation, the
  Agent still captures almost half of the aggregate performance mass
  (\(48.7\,\%\) of the total macro-F\textsubscript{1}), while Gemini
  accounts for \(36.1\,\%\) and OpenAI only \(15.2\,\%\).
  The result mirrors our tabular findings in
  Table~\ref{tab:results}: transparent, schema-aware reasoning yields
  the most faithful budget breakdowns.
  
  \paragraph{Evidence-extraction robustness (Fig.~\ref{fig:scatter-evidence}).}
  Across the vast majority of MDB projects the Agent achieves the highest
  F\textsubscript{1} (yellow), with Gemini (orange) and OpenAI (red)
  clustered below.
  An exception emerges for the grey-shaded projects, whose budgets are
  \emph{not} presented in explicit tables but scattered throughout the
  narrative text.
  Here Gemini’s end-to-end comprehension slightly outperforms the Agent,
  suggesting that large black-box models retain an advantage when
  numerical clues are deeply embedded in prose.

Taken together, the graphics align with our qualitative findings:
\emph{Glass-Box transparency dominates performance}—especially for structured or semi-structured financial disclosures.
While, black-box assistants narrow the gap only in the rare cases where budget
figures are diffused across free-form text.
Future work will therefore focus on augmenting the Agent’s retrieval
module with paragraph-level numerical parsing to close the remaining
gap on unstructured layouts.

\begin{figure*}[htp]
    \centering	\includegraphics[width = 1\textwidth]{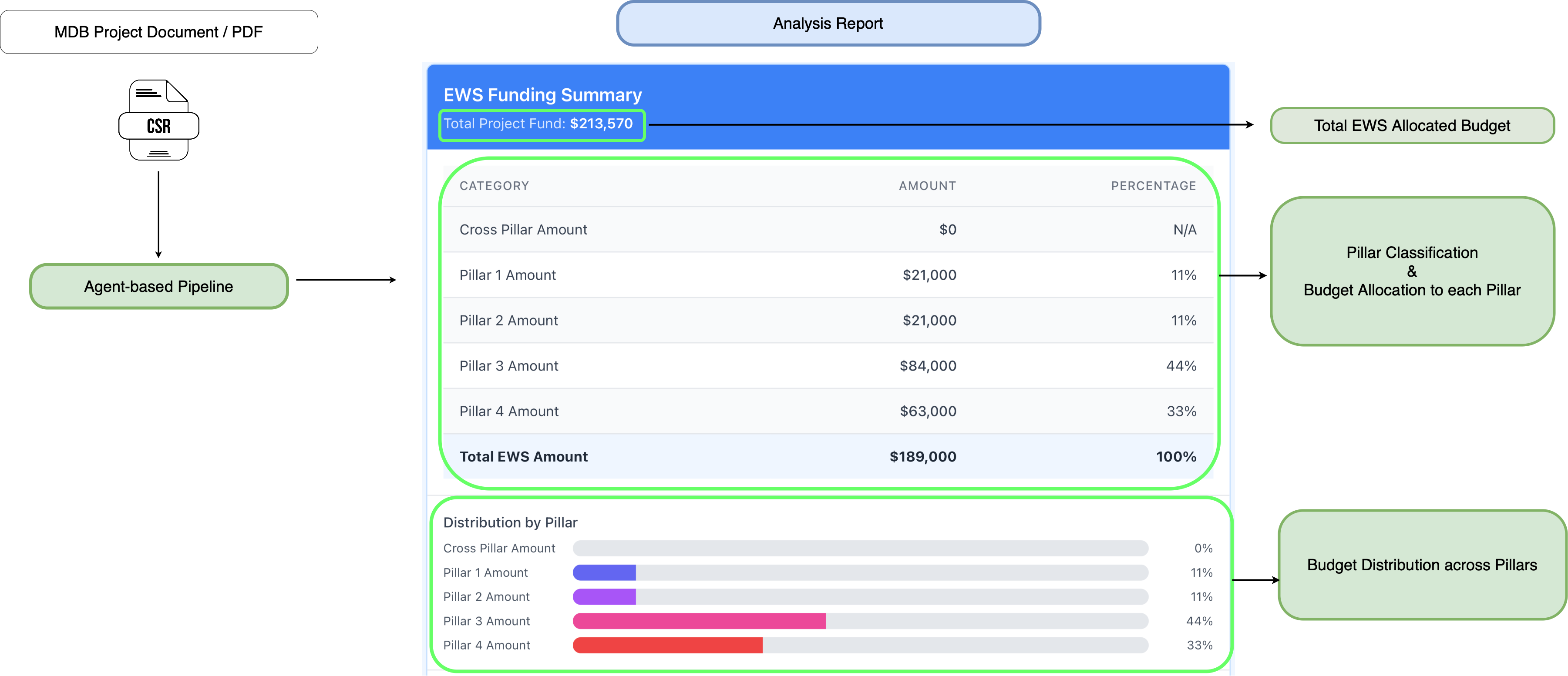}
    \vspace{-0.3em}
    \caption{Schematic overview of the final analysis report that results from the agent-based pipeline.The workflow comprises three main stages: (i) ingestion of the project document as a PDF; (ii) a modular agent-based processing pipeline that parses text, identifies total funding figures, and classifies expenditures into four predefined EWS “pillars”; and (iii) compilation of an analysis report summarizing the total allocated budget, per-pillar allocation amounts and percentages, and a graphical distribution of funds across pillars.}
    \label{fig:outupt1}
    \vspace{-0.7em}
    \end{figure*}

\section{Conclusion}
Automating financial tracking of EWS investments is crucial for improving climate finance transparency and accountability. In this study, we introduced the EWS4All Financial Tracking AI-Assistant (Fig.~\ref{fig:outupt}), a novel system that integrates multi-modal processing, hierarchical reasoning, and RAG for document classification and budget allocation.  Our experiments on 25 project documents from the CREWS Fund demonstrated that an agent-based approach significantly outperforms traditional NLP methods, achieving 87\% accuracy, 89\% precision, and 83\% recall. The system effectively addresses challenges related to document heterogeneity, structured and unstructured data integration, and cross-organizational inconsistencies. Beyond improving financial tracking, our work contributes a benchmark dataset for future AI research in climate finance. By combining AI-driven classification, retrieval, and reasoning, this approach enhances decision-making processes in MDBs and supports evidence-based climate investment policies.  Future work will focus on extending the system to handle a broader range of MDB financial documents, improving model generalization, and integrating real-time updates for dynamic financial tracking.

\section*{Limitations}
While our approach demonstrates significant improvements in automating financial tracking for EWS investments, several limitations remain. First, our system relies on existing financial reports from MDBs, in this case CREWS, which are often heterogeneous and may contain incomplete or ambiguous financial allocations. In cases where funding details are missing or inconsistently reported, even advanced retrieval-augmented generation (RAG) and multi-step reasoning approaches may struggle to provide accurate classifications. Second, the classification system is influenced by the training data used in fine-tuning and prompt engineering. Despite expert annotations, the model may still exhibit biases in investment classification, particularly when encountering novel financial structures or terminology not well-represented in the dataset. Third, while our agent-based RAG system achieves state-of-the-art performance on structured and unstructured financial data, its generalizability to other climate finance applications outside EWS has not been fully explored. Future work should assess model robustness across different sustainability reporting frameworks and financial instruments. Finally, our system assumes that financial tracking can be improved through AI-assisted reasoning; however, its real-world effectiveness depends on institutional adoption, policy integration, and alignment with evolving financial disclosure regulations.

\section*{Ethics Statement}

\myparagraph{Human Annotation}: This study relies on annotations provided by domain experts from the WMO, who possess extensive knowledge of Early Warning Systems (EWS). These experts played a pivotal role in the design and conceptualization of the study. Their deep understanding of both the contextual and practical aspects of the collected data ensures the accuracy and relevance of the annotations. The use of expert annotations minimizes the risk of misclassification and enhances the reliability of the model’s outputs.

\myparagraph{Responsible AI Use.}  
This tool is intended as an assistive system to enhance transparency and efficiency in financial tracking, not as a replacement for human analysts. Expert oversight remains crucial in interpreting financial classifications, addressing edge cases, and ensuring compliance with policy frameworks. By open-sourcing our dataset and model, we encourage responsible use and further validation to refine the system’s applicability in real-world climate finance decision-making.

\myparagraph{Data Privacy and Bias}: This study does not involve any personally identifiable or sensitive financial data. All data used in this research originates from publicly available sources under a Creative Commons license, ensuring compliance with data privacy regulations. While we find no evidence of demographic biases in the dataset, we acknowledge that financial reporting by multilateral development banks (MDBs) may reflect institutional biases in investment classification. Our model operates as a decision-support tool and should not replace human judgment in financial tracking and policy decisions.

\myparagraph{Reproducibility Statement}: To ensure full reproducibility, we will release all PDFs, codes, EWS-taxonomy, and expert-annotated data used in this study. Our approach aligns with best practices in AI transparency and responsible research dissemination. However, we encourage users of this dataset and model to consider ethical implications when applying automated financial tracking systems in real-world decision-making contexts. For vector database storage and retrieval, we utilized Weaviate, an open-source, scalable vector search engine that efficiently indexes high-dimensional embeddings. Additionally, for reasoning and large language model (LLM) interactions, we integrated OpenAI's o1 API, leveraging its advanced capabilities to process, analyze, and infer patterns from financial document data.

\ifarxiv
\section*{Disclaimer} 
Opinions expressed in this article are the author’s opinions and do not necessarily reflect those of WMO or its Members.

\section*{Acknowledgements} 
This paper has partially received funding from the Swiss
National Science Foundation (SNSF) under the project `How sustainable is sustainable finance? Impact evaluation and automated greenwashing detection' (Grant Agreement No. 100018\_207800).
\fi

\bibliography{custom1, additional_lit}

\appendix

\section{Early Warning Systems (EWS)\label{app:EWS}}

\subsection{Definition and Purpose}

Early Warning Systems (EWS) are integrated frameworks designed to detect imminent hazards and alert authorities and communities before disasters strike. In essence, an EWS combines hazard monitoring, risk analysis, communication, and preparedness planning to enable timely, preventive actions. Early warnings are a cornerstone of disaster risk reduction (DRR) – they save lives and reduce economic losses by giving people time to evacuate, protect assets, and secure critical infrastructure\footnote{See \url{https://www.unisdr.org/files/608_10340.pdf}.}. By empowering those at risk to act ahead of a hazard, EWS help build climate resilience: they are proven to safeguard lives, livelihoods, and ecosystems amid increasing climate-related threats\footnote{See, \url{https://www.unep.org/topics/climate-action/climate-transparency/climate-information-and-early-warning-systems}.}. In summary, an effective EWS ensures that impending dangers are rapidly identified, warnings reach the impacted population, and appropriate protective measures are taken in advance.

\subsection{EWS Taxonomy}

A robust EWS involves several fundamental components that work together seamlessly. The United Nations identify four interrelated pillars necessary for an effective people-centered EWS \cite{pescaroli2025bridging}. This taxonomy serves as a structured framework to categorize EWS components and activities, facilitating a consistent approach to analyzing early warning systems across various domains. Our approach in this paper is based on these four fundamental pillars of EWS and one cross-pillar, ensuring a comprehensive understanding of risk knowledge, detection, communication, and preparedness.

\begin{tcolorbox}[colback=gray!10, colframe=black, title=Early Warning System (EWS) Taxonomy Prompt, breakable]
An Early Warning System (EWS) is an integrated system of hazard monitoring, forecasting, and prediction, disaster risk assessment, communication, and preparedness activities that enables individuals, communities, governments, businesses, and others to take timely action to reduce disaster risks before hazardous events occur. 

When analyzing a text, it is essential to determine whether it falls under EWS components and activities, which vary across multiple sectors and require coordination and financing from various actors.

\textbf{The taxonomy is based on the Four Pillars of Early Warning Systems and one cross-pillar:}

\subsection*{Pillar 1: Disaster Risk Knowledge and Management (Led by UNDRR)}
This pillar focuses on understanding disaster risks and enhancing the knowledge of communities by collecting and utilizing comprehensive information on hazards, exposure, vulnerability, and capacity.

\textbf{Illustrative examples:}
\begin{itemize}[leftmargin=*, label={--}, nosep]
    \item Inclusive risk knowledge: Incorporating local, traditional, and scientific risk knowledge.
    \item Production of risk knowledge: Establishing a systematic recording of disaster loss data.
    \item Risk-informed planning: Ensuring decision-makers can access and use updated risk information.
    \item Data rescue: Digitizing and preserving historical disaster data.
\end{itemize}

\textbf{Keywords:} Risk mapping, vulnerability mapping, disaster risk reduction (DRR), climate information.

\hrulefill
\subsection*{Pillar 2: Detection, Observation, Monitoring, Analysis, and Forecasting (Led by WMO)}
This pillar enhances the capability to detect and monitor hazards, providing timely and accurate forecasting.

\textbf{Illustrative examples:}
\begin{itemize}[leftmargin=*, label={--}, nosep]
    \item Observing networks enhancement: Strengthening real-time monitoring systems.
    \item Hazard-specific observations: Improving monitoring of high-impact hazards.
    \item Impact-based forecasting: Developing quantitative triggers for anticipatory action.
\end{itemize}

\textbf{Keywords:} Forecasting, seasonal predictions, multi-model projections, climate services.

\hrulefill
\subsection*{Pillar 3: Warning Dissemination and Communication (Led by ITU)}
Effective communication ensures that early warnings are received by those at risk, enabling them to take timely action.

\textbf{Illustrative examples:}
\begin{itemize}[leftmargin=*, label={--}, nosep]
    \item Multichannel alert systems: Use of SMS, satellite, sirens, and social media.
    \item Standardized warnings: Implementation of the Common Alerting Protocol (CAP).
    \item Feedback mechanisms: Enabling community input on warning effectiveness.
\end{itemize}

\textbf{Keywords:} Communication systems, multichannel dissemination, emergency broadcast systems.

\hrulefill
\subsection*{Pillar 4: Preparedness and Response Capabilities (Led by IFRC)}
Timely preparedness and response measures translate early warnings into life-saving actions.

\textbf{Illustrative examples:}
\begin{itemize}[leftmargin=*, label={--}, nosep]
    \item Emergency preparedness planning: Developing anticipatory action frameworks.
    \item Public awareness campaigns: Educating communities on disaster response.
    \item Emergency shelters: Construction of cyclone shelters, evacuation centers.
\end{itemize}

\textbf{Keywords:} Preparedness planning, emergency drills, public education on disaster response.

\hrulefill
\subsection*{Cross-Pillar: Foundational Elements for Effective EWS}
Cross-cutting elements critical to the sustainability and effectiveness of EWS include governance, inclusion, institutional arrangements, and financial planning.

\textbf{Illustrative examples:}
\begin{itemize}[leftmargin=*, label={--}, nosep]
    \item Governance and institutional frameworks: Defining roles of agencies and stakeholders.
    \item Financial sustainability: Mobilizing and tracking finance for early warning systems.
    \item Regulatory support: Developing and enforcing data-sharing legislation.
\end{itemize}

\textbf{Keywords:} Institutional frameworks, governance, financial sustainability, data management.

\end{tcolorbox}

Each of these components is vital. Only when risk knowledge, monitoring, communication, and preparedness work in unison can an early warning system effectively protect lives and properties. Gaps in any one element (for example, if warnings don’t reach the vulnerable, or if communities don’t know how to respond) will weaken the whole system. Thus, successful EWS are people-centered and end-to-end, linking high-tech hazard detection with on-the-ground community action.

\subsection{Importance for climate finance} 
EWS are widely recognized as a high-impact, cost-effective investment for climate resilience. By providing advance notice of floods, storms, heatwaves and other climate-related hazards, EWS significantly reduce disaster losses. Studies indicate that every \$1 spent on early warnings can save up to \$10 by preventing damages and losses.\footnote{See, \url{https://wmo.int/news/media-centre/early-warnings-all-advances-new-challenges-emerge}.}  For example, just 24 hours’ warning of an extreme event can cut ensuing damage by about 30\%, and an estimated USD \$800 million investment in early warning infrastructure in developing countries could avert \$3–16 billion in losses every year\footnote{See, \url{https://www.unep.org/topics/climate-action/climate-transparency/climate-information-and-early-warning-systems}.}. These economic benefits underscore why EWS are considered “no-regret” adaptation measures, i.e., they pay for themselves many times over by protecting lives, assets, and development gains.

Given their proven value, EWS have become a priority in climate change adaptation and disaster risk reduction funding. International climate finance mechanisms, such as the Green Climate Fund, Climate Risk and Early Warning Systems (CREWS) Fund, and Adaptation Fund along with development banks, are channeling resources into EWS projects, from modernizing meteorological services and hazard monitoring networks to community training and alert communication systems. Strengthening EWS is also central to global initiatives like the United Nations’ Early Warnings for All (EW4All), which calls for expanding early warning coverage to 100\% of the global population by 2027. Achieving this goal requires substantial financial support to build new warning systems in climate-vulnerable countries and to maintain and upgrade existing ones. Climate finance is therefore being directed to help develop, implement, and sustain EWS, ensuring that countries can operate these systems (e.g. funding for equipment, data systems, and personnel) over the long term. In summary, investing in EWS is essential for climate resilience. It not only reduces humanitarian and economic impacts from extreme weather, but also yields high returns on investment. Financial support for EWS, whether through dedicated climate funds, loans and grants, or public budgets, underpins their development and sustainability, making it possible to deploy cutting-edge technology and foster prepared communities. By mitigating the worst effects of climate disasters, EWS help safeguard development progress, which is why they feature prominently in climate adaptation financing and strategies.

Hence, investing in EWS is essential for climate resilience. It not only reduces humanitarian and economic impacts from extreme weather, but also yields high returns on investment. Financial support for EWS, whether through dedicated climate funds, loans and grants, or public budgets, underpins their development and sustainability, making it possible to deploy cutting-edge technology and foster prepared communities. By mitigating the worst effects of climate disasters, EWS help safeguard development progress, which is why they feature prominently in climate adaptation financing and strategies.

\subsection{Current challenges}

Despite their clear benefits, there are several challenges in financing and implementing EWS effectively. Key issues include:

\paragraph{Data Inconsistencies and Lack of Standardization:} EWS rely on data from multiple sources (weather observations, risk databases, etc.), but often this data is inconsistent, incomplete, or not shared effectively across systems. Differences in how hazards are monitored and reported can lead to gaps or delays in warnings. Likewise, there is a lack of standardization in early warning protocols and data formats between agencies and countries \cite{velazquez2020review, pescaroli2025bridging}. Incompatible data systems and inconsistent methodologies (for example, different trigger criteria for warnings or varying risk assessment methods) make it difficult to integrate information. This fragmentation hinders the creation of a “common operating picture” of risk. Data harmonization and common standards (for data collection, forecasting models, and warning communication) are needed to ensure EWS components work together seamlessly.

\paragraph{Institutional and Cross-Organizational Barriers:} An effective EWS cuts across many organizations, national meteorological services, disaster management agencies, local governments, international partners, and communities. Coordinating these actors remains a challenge. In many cases, efforts are siloed: meteorological offices may issue technical warnings that don’t fully reach or engage local authorities or the public. There are gaps in governance, clarity of roles, and inter-agency communication that can weaken the warning chain. Improving EWS often requires overcoming bureaucratic boundaries and fostering cooperation between different sectors (e.g., linking climate scientists with emergency planners). Interoperability issues, i.e.,ensuring different organizations’ technologies and procedures align, are also a hurdle \cite{tupper2023mind}. As the World Meteorological Organization (WMO) states, connecting all relevant actors (from international agencies down to community groups) and adapting plans to real-world local conditions is complex\footnote{See, \url{https://wmo.int/news/media-centre/early-warnings-all-advances-new-challenges-emerge}.}. Sustained commitment, clear protocols, and partnerships are required to break down these barriers so that EWS operate as a cohesive, cross-sector system.

\paragraph{Financing Gaps and Sustainability:} While funding for EWS is rising, it still lags behind what is needed for global coverage and maintenance. Many high-risk developing countries lack the resources to install or upgrade EWS infrastructure (radar, sensors, communication tools) and to train personnel. Fragmented financing is a problem. Support comes from various donors and programs without a unified strategy, leading to potential overlaps in some areas and stark gaps in others. For instance, recent analyses show that a large share of EWS funding is concentrated in a few countries, while Small Island Developing States (SIDS) and Least Developed Countries (LDCs) remain underfunded despite being highly vulnerable\footnote{See, \url{https://wmo.int/media/news/tracking-funding-life-saving-early-warning-systems}.}. Even when initial capital is provided to set up an EWS, securing long-term funding for operations and maintenance (software updates, staffing, equipment calibration) is difficult. Without sustainable financing, systems can degrade over time. Ensuring financial sustainability, co-financing arrangements, and political commitment is critical so that EWS are not one-off projects but enduring services.

In addition to the above, there are challenges in technological adoption and last-mile delivery: for example, reaching remote or marginalized populations with warnings (issues of language, literacy, and reliable communication channels) and building trust so that people heed warnings. Climate change is also introducing new complexities – hazards are becoming more unpredictable or intense, testing the limits of existing early warning capabilities. Overall, addressing data and standardization issues, improving institutional coordination, and closing funding gaps are priority challenges to fully realize the life-saving potential of EWS.

\subsection{Relevance to this study}

Our work is focused on the financial tracking and classification of investments in climate resilience, and EWS represent a prime example of such investments. Early warning projects often cut across sectors and funding sources – they might include components of infrastructure, technology, capacity building, and community outreach. Because of this cross-cutting nature, tracking where and how money is spent on EWS can be difficult without a clear classification system. Different organizations may label EWS-related activities in various ways (e.g. “hydromet modernization”, “disaster preparedness”, “climate services”), leading to inconsistencies in investment data. By establishing a standardized framework to define and categorize EWS investments, the study helps create a “big-picture view” of early warning financing. This enables analysts and policymakers to identify overlaps, gaps, and trends that were previously obscured by fragmented data.

Moreover, improving the classification of EWS funding directly supports broader resilience initiatives. For instance, the newly launched Global Observatory for Early Warning System Investments is already working to tag and track EWS-related expenditures across major financial institutions. Such efforts mirror the goals of this study by highlighting the need for consistent tracking, transparency, and coordination in climate resilience finance. Better classification of investments means stakeholders can pinpoint where resources are going and where additional support is needed to meet global targets like the “Early Warnings for All by 2027” pledge. In short, EWS feature in this study as a critical category of climate resilience investment that must be clearly identified and monitored.

By including EWS in its financial tracking framework, the study provides valuable insights for decision-makers. It helps determine how much funding is allocated to early warnings, from which sources, and for what components (equipment, training, maintenance, etc.). This information is crucial for evidence-based decisions on scaling up EWS: for example, spotting a shortfall in community-level preparedness funding, or recognizing successful investment patterns that could be replicated. Ultimately, linking EWS to the study’s financial tracking reinforces the message that climate resilience investments can be better managed when we know their size, scope, and impact area. By classifying EWS expenditures systematically, the study contributes to stronger accountability and strategic planning in building climate resilience, ensuring that early warning systems – and the communities they protect – get the support they urgently need.

\section{Dataset Construction}
In this study, we analyze financial information extracted from PDFs containing both structured and unstructured data. Unlike conventional benchmark datasets, these documents exhibit high heterogeneity in their formats—some tables are well-structured, while others embed financial figures within free-text paragraphs or are scattered across multiple rows and columns. Additionally, many numerical values correspond to multiple rows within the same column, creating challenges in extraction, alignment, and interpretation.

The annotated data, provided by experts in CSV format, along with the corresponding PDFs, can be found in the supplementary materials of this paper.

The dataset consists of 298 rows of expert annotations and contains the following 9 columns:  
\textit{Fund, Project ID, Component, Outcome/Expected-Outcome/Objectives, Output/Sub-component, Activity/Output Indicator, Page Number, Amount,} and \textit{Label}.  

The total amount of Early Warning Systems (EWS) is computed as the sum of all \textit{Amount} values for a given project.

The annotated dataset (CSV file and PDFs) consists of financial reports and investment documents sourced from publicly available institutional records, which are intended for public information and research and transparency purposes. The dataset is used strictly within its intended scope—analyzing financial tracking in climate investments—and adheres to the original access conditions. Additionally, for the artifacts we create, including benchmark datasets and classification models, we specify their intended use for research and evaluation in automated financial tracking and ensure they remain compliant with ethical research guidelines.

\end{document}

\usepackage{enumitem}
\usepackage{array}
\usepackage{amsthm}
\usepackage{booktabs} 

\usepackage{listings}
\usepackage{longtable}
\usepackage{booktabs}
\usepackage{soul,xcolor}
\usepackage{longtable}
\sethlcolor{yellow}
\usepackage{subcaption}
\usepackage{multirow}
\usepackage{bm}
\usepackage{svg}
\newtheorem{definition}{Definition}
\usepackage{calligra}
\newcommand{\specialfont}[1]{{\itshape\calligra #1}}
\usepackage{tikz}
\newcommand{\grayb}{\textcolor{gray}{50.00}}
\newcommand{\grayz}{\textcolor{gray}{0.00}}
\definecolor{taborange}{rgb}{1.0, 0.498, 0.0549}
\definecolor{tabblue}{rgb}{0.1216, 0.4667, 0.7059}
\newcommand{\orangedotline}{%
    \tikz[baseline=-0.5ex]{
        \draw[thick, taborange] (0,0) -- (0.6,0);
        \filldraw[orange] (0.3,0) circle (0.07cm);
    }
}

\newcommand{\orangedottedline}{%
    \tikz[baseline=-0.5ex]{
        \draw[thick, taborange, dashed] (0,0) -- (0.6,0);
        \filldraw[orange] (0.3,0) circle (0.07cm);
    }
}

\newcommand{\bluexline}{%
    \tikz[baseline=-0.5ex]{
        \draw[thick, tabblue] (0,0) -- (0.6,0);
        \draw[thick, tabblue] (0.2,-0.08) -- (0.4,0.08);
        \draw[thick, tabblue] (0.4,-0.08) -- (0.2,0.08);
    }
}

\newcommand{\bluedottedxline}{%
    \tikz[baseline=-0.5ex]{
        \draw[thick, tabblue, dashed] (0,0) -- (0.6,0);
        \draw[thick, tabblue] (0.2,-0.08) -- (0.4,0.08);
        \draw[thick, tabblue] (0.4,-0.08) -- (0.2,0.08);
    }
}

\lstset{%
	basicstyle={\footnotesize\ttfamily},
	numbers=left,numberstyle=\footnotesize,xleftmargin=2em,
	aboveskip=1.5pt,belowskip=0pt,%
	showstringspaces=false,tabsize=2,breaklines=true}
 \usepackage[most]{tcolorbox}

\newtcolorbox{myquotebox}{
  colback=white!0, 
  colframe=black, 
  rounded corners,
  boxrule=0.5pt, 
  title=Prompt:,
  left=2mm, 
  right=2mm, 
  top=1mm, 
  bottom=1mm 
}

\usepackage{todonotes}
\usepackage{xspace}
\newcommand{\fixme}[2][]{\todo[color=goldenrod,size=\scriptsize,fancyline,caption={},#1]{#2}} 
\newcommand{\note}[4][]{\todo[author=#2,color=#3,size=\scriptsize,fancyline,caption={},#1]{#4}} 
\newcommand{\mrinmaya}[2][]{\note[#1]{mrinmaya}{green}{#2}\xspace}
\newcommand{\jingwei}[2][]{\note[#1]{jingwei}{yellow}{#2}\xspace}

\definecolor{lightgrey}{RGB}{158, 158, 158}
\definecolor{goldenrod}{rgb}{0,0,0.8}
\definecolor{deepred}{rgb}{0.6,0,0}
\definecolor{deepgreen}{rgb}{0,0.5,0}
\definecolor{pink}{RGB}{219, 48, 122}
\definecolor{forestgreen}{RGB}{34,139,34}
\definecolor{goldenrod}{RGB}{218,165,32}
\definecolor{sepia}{RGB}{112,66,20}

\usepackage[capitalise]{cleveref}
\crefname{figure}{Fig.}{Figs.}
\crefname{table}{Table}{Tables}
\crefname{appendix}{App.}{App.}
\crefname{section}{§}{§§}
\crefname{equation}{Eq.}{Eqs.}

\newcommand{\algname}{UNIT}

\newcommand{\smallgreen}[1]{\textcolor{deepgreen}{\scriptsize #1}}
\newcommand{\smallred}[1]{\textcolor{red}{\scriptsize #1}}
\definecolor{lightred}{RGB}{254, 138, 138}
\definecolor{lightblue}{RGB}{176, 195, 248}
\definecolor{lightgreen}{RGB}{138, 218, 174}
\newcommand\myparagraph[1]{
\vskip 0.05in 
\noindent{\bf {#1}}}
\newcommand*\samethanks[1][\value{footnote}]{\footnotemark[#1]}

\definecolor{boxborder}{RGB}{86, 113, 209}  
\definecolor{boxbg}{RGB}{255, 255, 255}    
\definecolor{boxtitle}{RGB}{255, 255, 255} 
\definecolor{boxheader}{RGB}{86, 113, 209}  

\makeatletter
\g@addto@macro\normalsize{%
  \setlength{\abovedisplayskip}{4pt}
  \setlength{\belowdisplayskip}{4pt}
  \setlength{\abovedisplayshortskip}{4pt}
  \setlength{\belowdisplayshortskip}{4pt}
}
\makeatother

\usepackage{tikz} 
\usepackage{xcolor} 

\newcommand{\bcircle}[1]{%
    \tikz[baseline=(char.base)]{
        \node[shape=circle,fill=black,draw,inner sep=1.5pt] (char) 
        {\textcolor{white}{\textbf{#1}}};}}

\title{AI for Climate Finance: Agentic Retrieval and Multi-Step Reasoning for Early Warning System Investments}

\author{ 
    \textbf{Saeid Ario Vaghefi}\textsuperscript{\rm 1,2}\thanks{Equal Contributions.},
    \textbf{Aymane Hachcham}\textsuperscript{\rm 1}\samethanks\\
    \textbf{Veronica Grasso}\textsuperscript{\rm 2},
    \textbf{Jiska Manicus}\textsuperscript{\rm 2}\\
    \textbf{Nakiete Msemo}\textsuperscript{\rm 2},
    \textbf{Chiara Colesanti Senni}\textsuperscript{\rm 1}\\
    \textbf{Markus Leippold}\textsuperscript{\rm 1,3} \\
    \textsuperscript{\rm 1}University of Zurich \hspace{5mm}
    \textsuperscript{\rm 2}WMO \hspace{5mm}
    \textsuperscript{\rm 3}Swiss Finance Institute (SFI) \\
    \texttt{\{saeid.vaghefi, aymane.hachcham, chiara.colesantisenni, markus.leippold\}@df.uzh.ch} \\
    \texttt{\{svaghefi, vgrasso, jmanicus, nmsemo\}@wmo.int}
}

\begin{document}
\maketitle
\begin{abstract}
Tracking financial investments in climate adaptation is a complex and expertise-intensive task, particularly for Early Warning Systems (EWS), 
which lack standardized financial reporting across multilateral development banks (MDBs) and funds. 
To address this challenge, we introduce an agent-based Retrieval-Augmented Generation (RAG) system that orchestrates contextual retrieval with internal chain-of-thought reasoning to extract relevant financial data, classify investments, and ensure compliance with funding guidelines. 
Our study focuses on a real-world application: tracking EWS investments in the Climate Risk and Early Warning Systems (CREWS) Fund. 
We evaluate our agent-based RAG pipeline on 25 MDB project documents from the CREWS Fund, 
comparing it against five model candidates—(1) a Zero-Shot Classifier (Baseline), (2) a Few-Shot “Zero Rule” Classifier, (3) a fine-tuned transformer-based classifier, and (4) a Few-Shot-V2 CoT+ICL classifier—across both multi-label classification and budget allocation tasks. 
Our agent-based RAG achieves 87\% accuracy, 89\% precision, and 83\% recall, significantly outperforming these benchmarks. 
We also benchmark it against the Gemini 2.0 Flash AI Assistant, setting the stage for a comparative study of Glass-Box Agents versus Black-Box Assistants to quantify the benefits of an agentic pipeline in transparency, explainability, and performance. 
Finally, we release a benchmark dataset and expert-annotated corpus to catalyze further research in AI-driven climate finance tracking.\footnote{We will open-source all code, LLM generations, and human annotations. This can foster further innovation and development in this important area, leading to even more sophisticated and effective tools for managing climate finance.}
\end{abstract}

\section{Introduction} \label{sec:introduction}
Recent advances in Large Language Models (LLMs) have transformed investment tracking, financial reporting, and compliance monitoring in climate finance. However, tracking financial flows and categorizing investments in Early Warning Systems (EWS) remains challenging due to the lack of standardized structures and terminologies in financial reports from Multilateral Development Banks (MDBs) and climate funds.  

\myparagraph{Motivation.}  
Early Warning Systems (EWS) are essential for disaster risk reduction and climate resilience. 
The United Nations (UN) has prioritized universal EWS access by 2027 through its Early Warnings for All (EW4All) initiative, 
emphasizing that timely warnings reduce economic losses and save lives. 
Studies show that 24 hours of advance warning can reduce damages by 30\%, while every dollar invested in early warning systems saves up to ten 
dollars in avoided losses\footnote{See Appendix \ref{app:EWS} for more on EWS.}. Despite their importance, 
EWS investments lack financial transparency, as MDB reports often fail to classify and track funding allocations systematically. The lack of standardized financial reporting for EWS investments by MDBs and funds creates inefficiencies and hinders effective resource allocation. 

In this work, we frame investment tracking as a multi-label classification task—each text or table snippet may belong to one or more of the CREWS Fund’s pillars—and, once labels are assigned, we automatically extract budget allocations with grounding evidence spans directly from the PDF. The resulting output is a structured JSON mapping each pillar to its supporting evidence and allocated funds, vastly reducing the time and expertise required for manual review.
To make our task concrete, we adopt the following pillar definitions:
\begin{itemize}
    \item \textbf{Pillar 1, Disaster risk knowledge:} Comprehensive information on hazards, exposure, vulnerability, and capacity—including the production, rescue, sharing, and application of risk data to inform early action.
    \item \textbf{Pillar 2, Hazard detection and forecasting:} Non-structural capacity‐building and structural infrastructure for multi-hazard monitoring, analysis, forecasting, and data management (e.g., observing networks, forecasting models, radars).
    \item \textbf{Pillar 3, Warning dissemination and communication:} Non-structural systems and structural platforms (cell-broadcast, sirens, SMS, social media, TV/radio, public address) that ensure timely, people-centered delivery of warnings to all at-risk groups.
    \item \textbf{Pillar 4, Preparedness to respond:} Non-structural planning and training (contingency, anticipatory action, public education) alongside structural shelters and resource centers that translate warnings into life-saving measures.
    \item \textbf{Cross-Pillar, Governance and sustainability:} Cross‐cutting institutional arrangements, policy frameworks, stakeholder coordination, and financial planning necessary to sustain and scale the four core pillars.
\end{itemize}
\myparagraph{Context.}  
EW4All underscores the need for financial transparency in climate adaptation: clear tracking of fund flows can improve project monitoring and reduce disaster losses. However, MDB financial reports present a highly heterogeneous mix of structured tables, free-form text, and institution-specific jargon, without standardized categorization or terminology.
Classical NLP approaches-e.g. fine-tuned transformer classifiers or rule-based table parsers-are brittle in this setting, requiring extensive labeled data to cover every layout variation and often failing to generalize across documents \cite{karpukhin-etal-2020-dense}, \cite{chen2020tabfactlargescaledatasettablebased}.
Even layout-aware transformers (LayoutLM \cite{10.1145/3394486.3403172}, Longformer \cite{beltagy2020longformerlongdocumenttransformer}) assume some consistency  in formatting or demand expensive layout annotations.

\begin{figure*}[htp]
\centering	\includegraphics[width = 1\textwidth]{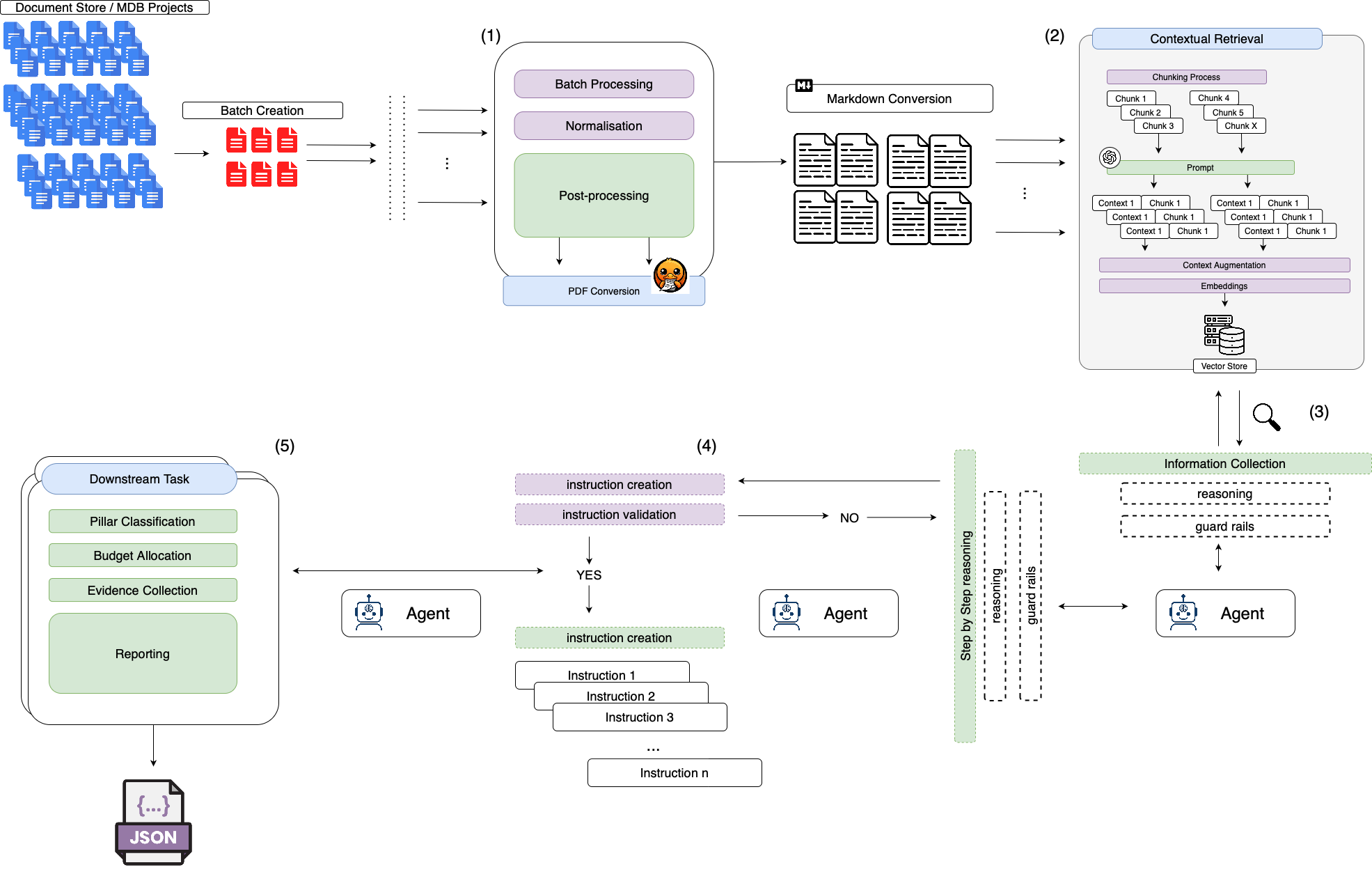}
\vspace{-0.3em}
\caption{AI-driven financial tracking pipeline for EWS investments. The different steps are: (1) PDF conversion, (2) context retrieval, (3) information storage and collection, (4) iterative sub-query and instruction creation, (5) dowstream task execution (pillar classification and budget allocation).}
\label{fig:pipeline}
\vspace{-0.7em}
\end{figure*}

To address these challenges, we argue that a multi-stage AI information system is essential.
By decomposing the task into dedicated components (c.f. Section~\ref{sec:methodology}, Figure~\ref{fig:pipeline}),
the pipeline can robustly handle diverse reporting formats, minimize annotation needs, and produce fully grounded, compact JSON outputs. 
This modular design leverages the strengths of each subcomponent to deliver the most reliable and scalable solution for climate finance transparency.

\myparagraph{Contribution.}  
We introduce the EW4All Financial Tracking AI-Assistant, an agent-based RAG pipeline that employs multi-modal extraction—parsing text, tables, and graphs—and internal chain-of-thought reasoning with built-in guardrails to produce robust, explainable decision chains across multiple sub-tasks.
We benchmark this approach against 4 model candidates—Zero-Shot Classifier (Baseline), Few-Shot “Zero Rule” Classifier, Fine-Tuned Transformer Classifier, and a Few-Shot-V2 CoT+ICL Classifier—on 25 CREWS-Fund documents, where it achieves 87\% accuracy, 89\% precision, and 83\% recall, 
a 23\% lift over traditional NLP methods.
We extend our evaluation to include the Gemini 2.0 Flash AI Assistant, setting up the first systematic contrast between transparent, agentic pipelines (Glass-Box Agents) and end-to-end black-box systems—quantifying gains in transparency, expert validation, and classification performance.
Finally, we open-source our expert-annotated corpus, benchmark dataset, and all prompt designs to catalyze future AI-driven climate finance tracking research.

\paragraph{Implications.}  
By improving climate finance transparency, this AI-driven approach provides structured, evidence-based insights into MDB investments. 
The integration of retrieval-augmented generation and agentic AI enhances decision-making, financial accountability, and policy formulation in global climate investment tracking. 
With a clearer understanding of investment patterns, gaps, and overlaps, stakeholders can make more informed decisions regarding resource allocation, project prioritization, and policy formulation in global climate investment tracking. 
The integration of retrieval-augmented generation (RAG) and agentic AI also enhances explainability and expert validation, making the system's outputs more reliable for decision-making. 
The evidence-based insights provided by the AI system can support the formulation of more effective climate adaptation policies. By identifying areas where investments are lacking or where funding guidelines might need adjustments, policymakers can use this information to optimize resource allocation for climate resilience. Hence, this work contributes to broader AI applications in climate finance, supporting international initiatives that seek to optimize resource allocation for climate resilience.

\section{Related Literature}





RAG improves knowledge-intensive tasks by integrating external retrieval with LLM generation \cite{lewis2020retrieval}, yet traditional RAG remains limited by static retrieval pipelines. Agentic RAG enhances adaptability by incorporating iterative retrieval and decision-making, improving factual accuracy and multi-step reasoning \cite{xi2023risepotentiallargelanguage, yao2023react, guo2024largelanguagemodelbased}. Multi-agent frameworks extend this by refining retrieval for applications such as code generation and verification \cite{guo2024largelanguagemodelbased, liu2024largelanguagemodelbasedagents}, advancing explainability and human-AI collaboration.  

In-Context Learning (ICL) allows LLMs to generalize from few-shot demonstrations without fine-tuning \cite{brown2020language}, but its effectiveness hinges on example selection. Retrieval-based ICL improves prompt efficiency, and reward models further refine in-context retrieval \cite{wang2024reinforcement}. CoT prompting facilitates step-by-step reasoning, significantly boosting performance in arithmetic and commonsense tasks \cite{wei2022chain, kojima2022large}. Self-consistency decoding enhances CoT by aggregating multiple reasoning paths \cite{wang2023selfconsistency}, while example-based prompting strengthens complex question-answering capabilities \cite{diao2024activepromptingchainofthoughtlarge}.


\section{Methodology}
\label{sec:methodology}
MDB project documents are characterized by highly heterogeneous layouts—mixed narrative text, nested tables, multi-column formats, footnotes, and embedded figures-such
that evidence of EWS pillars and funding may be dispersed across pages, tables, and descriptive passages.
Conventional retrieval or single-pass parsing pipelines struggle to (i) locate semantically related spans when they reside in separate structural regions,
(ii) reconcile duplicate or overlapping budget figures across distinct table formats and (iii) ensure end-to-end consistency in the face of OCR errors or layout ambiguities.

To address these challenges, we adopt an agent-based retrieval-augmented generation (RAG) framework that orchestrates:
\begin{enumerate}
    \item \textit{Iterative sub-query generation}, where an LLM-driven agent dynamically decomposes the overall extraction task into fine-grained retrieval instructions.
    \item \textit{Hybrid semantic-lexical search}, combining dense vector retrieval with BM25-style keyword matching to capture both contextual relevance and exact matches.
    \item \textit{Self-validation loops or guardrails}, in which the agent examines the sufficiency and coherence of retrieved chunks (re-issuing queries when coverage thresholds are unmet).
    \item \textit{Schema-aware consolidation}, formatting the final evidence spans and associated numeric allocations into a single structured JSON output.
\end{enumerate}

Figure~\ref{fig:pipeline} illustrates the overall pipeline with all its components.

\subsection{Embedding Construction and Indexing}

Effective downstream reasoning over MDB PDFs requires a robust embedding index that reconciles heterogeneous layouts and scattered evidence. 
To this end, we employ a unified four‐stage pipeline that breaks the task into four main components:  
document-parsing, chunking, context augmentation, embedding generation and vector storage. 
This is essential to capture the structural semantics and ensure retrieval fidelity.

First, we extract both raw text and structural elements from each document \(d\) using \textit{Docling} document converter \cite{auer2024doclingtechnicalreport}:
\[
T_{d} = \mathrm{DoclingParser}(d),
\]
and partition \(T_{d}\) into three disjoint chunk sets:
\[
\mathcal{C}_{\mathrm{table}},\quad \mathcal{C}_{\mathrm{text}},\quad \mathcal{C}_{\mathrm{image}},
\]
where \(\mathcal{C}_{\mathrm{table}}\) comprises automatically detected table regions, \(\mathcal{C}_{\mathrm{text}}\) contains narrative passages split at markdown‐style headers, and \(\mathcal{C}_{\mathrm{image}}\) includes embedded figures. Writing \(\mathcal{C}=\mathcal{C}_{\mathrm{table}}\cup\mathcal{C}_{\mathrm{text}}\cup\mathcal{C}_{\mathrm{image}}\), this decomposition prevents loss of context and mitigates parsing errors arising from complex multi‐column layouts and mixed content.

Next, to situate each chunk within its document context and reduce semantic ambiguity \cite{günther2024latechunkingcontextualchunk}, 
we generate a concise, two-sentence summary for each \(c\in\mathcal{C}\). We prompt an LLM with \(P_{\mathrm{ctx}}(c,T_{d})\) to obtain
\[
\mathrm{ctx}(c)\;=\;\mathrm{LLM}\bigl(P_{\mathrm{ctx}}(c,T_{d})\bigr),
\]
and form the augmented chunk
\[
c' = c \;\oplus\; \mathrm{ctx}(c).
\]
By anchoring each chunk to its global narrative, we ensure that subsequent retrieval captures both fine‐grained detail and overall document significance.

We encode each augmented chunk \(c'\) into two modality‐specific latent spaces: one jointly for text and tables, and one for images. Formally, we define  

\[
e_{\mathrm{tt}}(c') = f_{\mathrm{tt}}(c') \in \mathbb{R}^{d_{\mathrm{tt}}},\quad
e_{\mathrm{im}}(c') = f_{\mathrm{im}}(c') \in \mathbb{R}^{d_{\mathrm{im}}}
\]

where \(f_{t}\) is a joint text-table encoder trained to capture both narrative and structured tabular semantics, and \(f_{im}\) is an image encoder \cite{radford2021learningtransferablevisualmodels} 
specialized for visual feature extraction. 
We index these two embedding spaces in Weaviate by defining separate NamedVector configurations—one for text-table properties and another for image properties—thus 
preserving modality-specific semantics and enabling efficient hybrid (semantic + lexical) search across modalities. 
At query time, Weaviate dispatches each multimodal query to the appropriate vector index and returns the top-$k$ relevant chunks for downstream RAG orchestration.

This embedding step condenses high-dimensional text and layout features into a semantic space where related content remains in proximity.

Finally, each embedding \(e(c')\) is stored in a vector database with metadata \(\mathrm{meta}(c')=\{\texttt{file\_name}:f\}\) (where \(f\) is the PDF’s filename) by
\[
\mathrm{VDB\_store}\bigl(e(c'),\,\mathrm{meta}(c')\bigr).
\]
At inference time, for a given file ID \(f\) and query \(q\), we retrieve the top-5 semantically and lexically relevant chunks via
\[
\mathcal{R}(f) \;=\; \mathrm{VDB\_query}(q,\,f), 
\quad |\mathcal{R}(f)|=5,
\]

\section{Hybrid Retrieval via Rank Fusion}

In addition to the above procedure, we employ a hybrid search strategy that combines dense vector search with BM25F-based keyword search \cite{robertson2009probabilistic} to leverage both semantic similarity and exact lexical matching. Let $\mathcal{R}_v(q, f)$
denote the set of candidate chunks retrieved via dense vector search, and let $\mathcal{R}_k(q, f)$ denote the candidate chunks obtained via BM25F keyword search. To fuse these two retrieval sets, we use Reciprocal Rank Fusion (RRF) \cite{cormack2009reciprocal}. For each candidate chunk \( c \in \mathcal{R}_v(q, f) \cup \mathcal{R}_k(q, f) \) we compute an RRF score as:
\[
\text{RRF}(c) = \sum_{i\in\{v,k\}} \frac{1}{\text{rank}_i(c) + K},
\]
where \( \text{rank}_i(c) \) is the rank of \( c \) in retrieval system \( i \) (with lower ranks corresponding to higher relevance) and \( K \) is a smoothing constant (typically set to 60). The final set of retrieved chunks is then given by selecting the top five candidates according to their RRF scores:
\[
\mathcal{R}(f) = \operatorname{Top5}\Bigl(\mathcal{R}_v(q, f) \cup \mathcal{R}_k(q, f), \, \text{RRF}(c)\Bigr).
\]
This hybrid method harnesses the semantic sensitivity of dense vector retrieval alongside the precise lexical matching of BM25F, thereby enhancing the overall disambiguation and retrieval performance during downstream processing.

\subsection{Classification and Budget Allocation}

For each retrieved chunk \( c' \in \mathcal{R}(f) \), we apply the following four methods to classify the chunk (i.e., assign it a class \( y \) from the five pillars) and to allocate an associated budget \( B \).

\subsubsection{Zero-Shot and Few-Shot Classification}

In this approach, we construct a prompt \( P_{\text{Class+Budget}}(c') \) that includes the content of the augmented chunk and (in the few-shot setting) several annotated examples. The LLM is then queried to simultaneously produce an outcome classification \( y \) and an associated budget \( B \):
\[
\{y, B\} = \text{LLM}(P_{\text{Class+Budget}}(c')).
\]
This method leverages the pre-trained knowledge of the LLM, with few-shot prompting guiding its responses.

\subsubsection{Fine-Tuned Transformer-Based Classifier}

In another approach, we fine-tune a transformer-based classifier \( M_{\text{ft}} \) on a labeled dataset \( \{(c'_i, y_i)\}_{i=1}^{N} \). The model is used to classify each augmented chunk $y = M_{\text{ft}}(c')$. 
Subsequently, an LLM is used to determine the budget allocation of each class. The prompt \( P_{\text{Budget}}(c', y) \) is constructed using the the chunk and its classification.
\[
B = \text{LLM}(P_{\text{Budget}}(c', y)).
\]
The final result for each chunk is the tuple \(\{y, B\}\).

\subsubsection{Few-Shot-V2: Chain-of-Thought (CoT)}

This approach employs a three-step COT strategy, resulting in a tuple \(\{y, B\}\):
\begin{enumerate}[label=\arabic*, nosep]
    \item \textbf{Reformatting:} If \( c' \) represents a table, it is reformatted into a clean markdown table:
    \[
    c'' = \text{LLM}(P_{\text{reformat}}(c')).
    \]
    Otherwise, we set \( c'' = c' \).
    \item \textbf{Classification:} A classification prompt is used to classify the (reformatted) chunk:
    \[
    y = \text{LLM}(P_{\text{Class}}(c'')).
    \]
    \item \textbf{Budget Allocation:} A subsequent prompt allocates the budget
    $B = \text{LLM}(P_{\text{Budget}}(c'', y))$.
\end{enumerate}

\subsubsection{Agent-Based Approach}

This method uses an agent that follows a sequence of instructions and performs RAG queries:
\begin{enumerate}[label=\arabic*., nosep]
    \item \textbf{Instruction Generation:} The agent, primed with examples of annotated PDFs and the desired output format, generates a list of sub-task instructions \( I = \{i_1, i_2, \ldots, i_k\} \) to complete the classification and budget allocation task. It also generates a list of queries \( Q = \{q_1, q_2, \ldots, q_l\} \) to use if the sub-tasks require querying the vdb.

    \item \textbf{Sub-Task and Query Mapping:} The agent maps instructions \( I\) to queries \( Q\).

    \item \textbf{Sub-Task Execution:} For each instruction \( i_j \), is the sub-task requires querying the vdb, a retrieval is performed to extract relevant chunks:
    \[
    c'_{i_j} = \text{VDB\_query}(q_{i_j}, f)
    \]

    \item \textbf{Sub-Task Validation:} The agent performs a self-healing step to validate that the retrieved chunks \(c'_{i_j}\) are sufficient. If not, a new query \(q_{i_j}^{\text{new}}\) is generated and the retrieval is repeated:
{\small
\[
c'_{i_j}{}^{\text{final}} =
\begin{cases}
\text{VDB\_query}(q_{i_j}^{\text{new}}, f), & \text{if } c'_{i_j} \text{ is insufficient}, \\[1ex]
c'_{i_j}, & \text{otherwise}.
\end{cases}
\]
}
 \item \textbf{Final Formatting:} After finishing all the sub-tasks, the final step formats the output as JSON:
    \[
    \{y, B\} = \text{LLM}(P_{\text{Format}}(\{\text{result}_{I}\}))
    \]
\end{enumerate}


\section{Results}
We evaluated our methodology on an evaluation set comprising a collection of PDF documents from the CREWS Fund. Our evaluation focuses on how accurately the budget is distributed across the EWS Pillars for each document. To this end, we assess three key metrics: accuracy, precision, and recall. Table~\ref{tab:results} summarizes the performance of each method, where the metrics for the agent-based approach are highlighted in bold due to its superior performance.

\begin{figure*}[htp]
    \centering	\includegraphics[width = 1\textwidth]{output.png}
    \vspace{-0.3em}
    \caption{Schematic overview of the final analysis report that results from the agent-based pipeline.The workflow comprises three main stages: (i) ingestion of the project document as a PDF; (ii) a modular agent-based processing pipeline that parses text, identifies total funding figures, and classifies expenditures into four predefined EWS “pillars”; and (iii) compilation of an analysis report summarizing the total allocated budget, per-pillar allocation amounts and percentages, and a graphical distribution of funds across pillars.}
    \label{fig:outupt}
    \vspace{-0.7em}
    \end{figure*}

\begin{table}[ht]
\centering
\resizebox{\columnwidth}{!}{%
\begin{tabular}{lccc}
\toprule
\textbf{Method} & \textbf{Accuracy} & \textbf{Precision} & \textbf{Recall} \\
\midrule
Zero-Shot       & 0.41 & 0.40 & 0.61 \\
Few-Shot        & 0.42 & 0.45 & 0.64 \\
Transformer     & 0.41 & 0.64 & 0.32 \\
Few-Shot-CoT    & 0.51 & 0.63 & 0.71 \\
Agent           & \textbf{0.87} & \textbf{0.89} & \textbf{0.83} \\
\bottomrule
\end{tabular}
}
\caption{Evaluation metrics for budget distribution across the EWS Pillars.}
\label{tab:results}
\end{table}
The results indicate that the agent-based approach significantly outperforms the other methods, achieving higher accuracy, precision, and recall. This suggests that the integration of retrieval-augmented generation and dynamic sub-task execution in the agent method greatly enhances the effectiveness of budget allocation across the pillars.

\section{Conclusion}
Automating financial tracking of EWS investments is crucial for improving climate finance transparency and accountability. In this study, we introduced the EWS4All Financial Tracking AI-Assistant, a novel system that integrates multi-modal processing, hierarchical reasoning, and RAG for document classification and budget allocation.  Our experiments on 25 project documents from the CREWS Fund demonstrated that an agent-based approach significantly outperforms traditional NLP methods, achieving 87\% accuracy, 89\% precision, and 83\% recall. The system effectively addresses challenges related to document heterogeneity, structured and unstructured data integration, and cross-organizational inconsistencies. Beyond improving financial tracking, our work contributes a benchmark dataset for future AI research in climate finance. By combining AI-driven classification, retrieval, and reasoning, this approach enhances decision-making processes in MDBs and supports evidence-based climate investment policies.  Future work will focus on extending the system to handle a broader range of MDB financial documents, improving model generalization, and integrating real-time updates for dynamic financial tracking.

\section*{Limitations}
While our approach demonstrates significant improvements in automating financial tracking for EWS investments, several limitations remain. First, our system relies on existing financial reports from MDBs, in this case CREWS, which are often heterogeneous and may contain incomplete or ambiguous financial allocations. In cases where funding details are missing or inconsistently reported, even advanced retrieval-augmented generation (RAG) and multi-step reasoning approaches may struggle to provide accurate classifications. Second, the classification system is influenced by the training data used in fine-tuning and prompt engineering. Despite expert annotations, the model may still exhibit biases in investment classification, particularly when encountering novel financial structures or terminology not well-represented in the dataset. Third, while our agent-based RAG system achieves state-of-the-art performance on structured and unstructured financial data, its generalizability to other climate finance applications outside EWS has not been fully explored. Future work should assess model robustness across different sustainability reporting frameworks and financial instruments. Finally, our system assumes that financial tracking can be improved through AI-assisted reasoning; however, its real-world effectiveness depends on institutional adoption, policy integration, and alignment with evolving financial disclosure regulations.

\section*{Ethics Statement}

\myparagraph{Human Annotation}: This study relies on annotations provided by domain experts from the World Meteorological Organization (WMO), who possess extensive knowledge of Early Warning Systems (EWS). These experts played a pivotal role in the design and conceptualization of the study. Their deep understanding of both the contextual and practical aspects of the collected data ensures the accuracy and relevance of the annotations. The use of expert annotations minimizes the risk of misclassification and enhances the reliability of the model’s outputs.

\myparagraph{Responsible AI Use.}  
This tool is intended as an assistive system to enhance transparency and efficiency in financial tracking, not as a replacement for human analysts. Expert oversight remains crucial in interpreting financial classifications, addressing edge cases, and ensuring compliance with policy frameworks. By open-sourcing our dataset and model, we encourage responsible use and further validation to refine the system’s applicability in real-world climate finance decision-making.

\myparagraph{Data Privacy and Bias}: This study does not involve any personally identifiable or sensitive financial data. All data used in this research originates from publicly available sources under a Creative Commons license, ensuring compliance with data privacy regulations. While we find no evidence of demographic biases in the dataset, we acknowledge that financial reporting by multilateral development banks (MDBs) may reflect institutional biases in investment classification. Our model operates as a decision-support tool and should not replace human judgment in financial tracking and policy decisions.

\myparagraph{Reproducibility Statement}: To ensure full reproducibility, we will release all PDFs, codes, EWS-taxonomy, and expert-annotated data used in this study. Our approach aligns with best practices in AI transparency and responsible research dissemination. However, we encourage users of this dataset and model to consider ethical implications when applying automated financial tracking systems in real-world decision-making contexts. For vector database storage and retrieval, we utilized Weaviate, an open-source, scalable vector search engine that efficiently indexes high-dimensional embeddings. Additionally, for reasoning and large language model (LLM) interactions, we integrated OpenAI's o1 API, leveraging its advanced capabilities to process, analyze, and infer patterns from financial document data.

\ifarxiv
\section*{Acknowledgements} 
This paper has received funding from the Swiss
National Science Foundation (SNSF) under the project `How sustainable is sustainable finance? Impact evaluation and automated greenwashing detection' (Grant Agreement No. 100018\_207800).
\fi

\bibliography{custom1, additional_lit}

\appendix

\section{Early Warning Systems (EWS)\label{app:EWS}}

\subsection{Definition and Purpose}

Early Warning Systems (EWS) are integrated frameworks designed to detect imminent hazards and alert authorities and communities before disasters strike. In essence, an EWS combines hazard monitoring, risk analysis, communication, and preparedness planning to enable timely, preventive actions. Early warnings are a cornerstone of disaster risk reduction (DRR) – they save lives and reduce economic losses by giving people time to evacuate, protect assets, and secure critical infrastructure\footnote{See \url{https://www.unisdr.org/files/608_10340.pdf}.}. By empowering those at risk to act ahead of a hazard, EWS help build climate resilience: they are proven to safeguard lives, livelihoods, and ecosystems amid increasing climate-related threats\footnote{See, \url{https://www.unep.org/topics/climate-action/climate-transparency/climate-information-and-early-warning-systems}.}. In summary, an effective EWS ensures that impending dangers are rapidly identified, warnings reach the impacted population, and appropriate protective measures are taken in advance.

\subsection{EWS Taxonomy}

A robust EWS involves several fundamental components that work together seamlessly. The United Nations identify four interrelated pillars necessary for an effective people-centered EWS \cite{pescaroli2025bridging}. This taxonomy serves as a structured framework to categorize EWS components and activities, facilitating a consistent approach to analyzing early warning systems across various domains. Our approach in this paper is based on these four fundamental pillars of EWS and one cross-pillar, ensuring a comprehensive understanding of risk knowledge, detection, communication, and preparedness.

\begin{tcolorbox}[colback=gray!10, colframe=black, title=Early Warning System (EWS) Taxonomy Prompt, breakable]
An Early Warning System (EWS) is an integrated system of hazard monitoring, forecasting, and prediction, disaster risk assessment, communication, and preparedness activities that enables individuals, communities, governments, businesses, and others to take timely action to reduce disaster risks before hazardous events occur. 

When analyzing a text, it is essential to determine whether it falls under EWS components and activities, which vary across multiple sectors and require coordination and financing from various actors.

\textbf{The taxonomy is based on the Four Pillars of Early Warning Systems and one cross-pillar:}

\subsection*{Pillar 1: Disaster Risk Knowledge and Management (Led by UNDRR)}
This pillar focuses on understanding disaster risks and enhancing the knowledge of communities by collecting and utilizing comprehensive information on hazards, exposure, vulnerability, and capacity.

\textbf{Illustrative examples:}
\begin{itemize}[leftmargin=*, label={--}, nosep]
    \item Inclusive risk knowledge: Incorporating local, traditional, and scientific risk knowledge.
    \item Production of risk knowledge: Establishing a systematic recording of disaster loss data.
    \item Risk-informed planning: Ensuring decision-makers can access and use updated risk information.
    \item Data rescue: Digitizing and preserving historical disaster data.
\end{itemize}

\textbf{Keywords:} Risk mapping, vulnerability mapping, disaster risk reduction (DRR), climate information.

\hrulefill
\subsection*{Pillar 2: Detection, Observation, Monitoring, Analysis, and Forecasting (Led by WMO)}
This pillar enhances the capability to detect and monitor hazards, providing timely and accurate forecasting.

\textbf{Illustrative examples:}
\begin{itemize}[leftmargin=*, label={--}, nosep]
    \item Observing networks enhancement: Strengthening real-time monitoring systems.
    \item Hazard-specific observations: Improving monitoring of high-impact hazards.
    \item Impact-based forecasting: Developing quantitative triggers for anticipatory action.
\end{itemize}

\textbf{Keywords:} Forecasting, seasonal predictions, multi-model projections, climate services.

\hrulefill
\subsection*{Pillar 3: Warning Dissemination and Communication (Led by ITU)}
Effective communication ensures that early warnings are received by those at risk, enabling them to take timely action.

\textbf{Illustrative examples:}
\begin{itemize}[leftmargin=*, label={--}, nosep]
    \item Multichannel alert systems: Use of SMS, satellite, sirens, and social media.
    \item Standardized warnings: Implementation of the Common Alerting Protocol (CAP).
    \item Feedback mechanisms: Enabling community input on warning effectiveness.
\end{itemize}

\textbf{Keywords:} Communication systems, multichannel dissemination, emergency broadcast systems.

\hrulefill
\subsection*{Pillar 4: Preparedness and Response Capabilities (Led by IFRC)}
Timely preparedness and response measures translate early warnings into life-saving actions.

\textbf{Illustrative examples:}
\begin{itemize}[leftmargin=*, label={--}, nosep]
    \item Emergency preparedness planning: Developing anticipatory action frameworks.
    \item Public awareness campaigns: Educating communities on disaster response.
    \item Emergency shelters: Construction of cyclone shelters, evacuation centers.
\end{itemize}

\textbf{Keywords:} Preparedness planning, emergency drills, public education on disaster response.

\hrulefill
\subsection*{Cross-Pillar: Foundational Elements for Effective EWS}
Cross-cutting elements critical to the sustainability and effectiveness of EWS include governance, inclusion, institutional arrangements, and financial planning.

\textbf{Illustrative examples:}
\begin{itemize}[leftmargin=*, label={--}, nosep]
    \item Governance and institutional frameworks: Defining roles of agencies and stakeholders.
    \item Financial sustainability: Mobilizing and tracking finance for early warning systems.
    \item Regulatory support: Developing and enforcing data-sharing legislation.
\end{itemize}

\textbf{Keywords:} Institutional frameworks, governance, financial sustainability, data management.

\end{tcolorbox}

Each of these components is vital. Only when risk knowledge, monitoring, communication, and preparedness work in unison can an early warning system effectively protect lives and properties. Gaps in any one element (for example, if warnings don’t reach the vulnerable, or if communities don’t know how to respond) will weaken the whole system. Thus, successful EWS are people-centered and end-to-end, linking high-tech hazard detection with on-the-ground community action.

\subsection{Importance for climate finance} 
EWS are widely recognized as a high-impact, cost-effective investment for climate resilience. By providing advance notice of floods, storms, heatwaves and other climate-related hazards, EWS significantly reduce disaster losses. Studies indicate that every \$1 spent on early warnings can save up to \$10 by preventing damages and losses.\footnote{See, \url{https://wmo.int/news/media-centre/early-warnings-all-advances-new-challenges-emerge}.}  For example, just 24 hours’ warning of an extreme event can cut ensuing damage by about 30\%, and an estimated USD \$800 million investment in early warning infrastructure in developing countries could avert \$3–16 billion in losses every year\footnote{See, \url{https://www.unep.org/topics/climate-action/climate-transparency/climate-information-and-early-warning-systems}.}. These economic benefits underscore why EWS are considered “no-regret” adaptation measures, i.e., they pay for themselves many times over by protecting lives, assets, and development gains.

Given their proven value, EWS have become a priority in climate change adaptation and disaster risk reduction funding. International climate finance mechanisms, such as the Green Climate Fund, Climate Risk and Early Warning Systems (CREWS) Fund, and Adaptation Fund along with development banks, are channeling resources into EWS projects, from modernizing meteorological services and hazard monitoring networks to community training and alert communication systems. Strengthening EWS is also central to global initiatives like the United Nations’ Early Warnings for All (EW4All), which calls for expanding early warning coverage to 100\% of the global population by 2027. Achieving this goal requires substantial financial support to build new warning systems in climate-vulnerable countries and to maintain and upgrade existing ones. Climate finance is therefore being directed to help develop, implement, and sustain EWS, ensuring that countries can operate these systems (e.g. funding for equipment, data systems, and personnel) over the long term. In summary, investing in EWS is essential for climate resilience. It not only reduces humanitarian and economic impacts from extreme weather, but also yields high returns on investment. Financial support for EWS, whether through dedicated climate funds, loans and grants, or public budgets, underpins their development and sustainability, making it possible to deploy cutting-edge technology and foster prepared communities. By mitigating the worst effects of climate disasters, EWS help safeguard development progress, which is why they feature prominently in climate adaptation financing and strategies.

Hence, investing in EWS is essential for climate resilience. It not only reduces humanitarian and economic impacts from extreme weather, but also yields high returns on investment. Financial support for EWS, whether through dedicated climate funds, loans and grants, or public budgets, underpins their development and sustainability, making it possible to deploy cutting-edge technology and foster prepared communities. By mitigating the worst effects of climate disasters, EWS help safeguard development progress, which is why they feature prominently in climate adaptation financing and strategies.

\subsection{Current challenges}

Despite their clear benefits, there are several challenges in financing and implementing EWS effectively. Key issues include:

\paragraph{Data Inconsistencies and Lack of Standardization:} EWS rely on data from multiple sources (weather observations, risk databases, etc.), but often this data is inconsistent, incomplete, or not shared effectively across systems. Differences in how hazards are monitored and reported can lead to gaps or delays in warnings. Likewise, there is a lack of standardization in early warning protocols and data formats between agencies and countries \cite{velazquez2020review, pescaroli2025bridging}. Incompatible data systems and inconsistent methodologies (for example, different trigger criteria for warnings or varying risk assessment methods) make it difficult to integrate information. This fragmentation hinders the creation of a “common operating picture” of risk. Data harmonization and common standards (for data collection, forecasting models, and warning communication) are needed to ensure EWS components work together seamlessly.

\paragraph{Institutional and Cross-Organizational Barriers:} An effective EWS cuts across many organizations, national meteorological services, disaster management agencies, local governments, international partners, and communities. Coordinating these actors remains a challenge. In many cases, efforts are siloed: meteorological offices may issue technical warnings that don’t fully reach or engage local authorities or the public. There are gaps in governance, clarity of roles, and inter-agency communication that can weaken the warning chain. Improving EWS often requires overcoming bureaucratic boundaries and fostering cooperation between different sectors (e.g., linking climate scientists with emergency planners). Interoperability issues, i.e.,ensuring different organizations’ technologies and procedures align, are also a hurdle \cite{tupper2023mind}. As the World Meteorological Organization (WMO) states, connecting all relevant actors (from international agencies down to community groups) and adapting plans to real-world local conditions is complex\footnote{See, \url{https://wmo.int/news/media-centre/early-warnings-all-advances-new-challenges-emerge}.}. Sustained commitment, clear protocols, and partnerships are required to break down these barriers so that EWS operate as a cohesive, cross-sector system.

\paragraph{Financing Gaps and Sustainability:} While funding for EWS is rising, it still lags behind what is needed for global coverage and maintenance. Many high-risk developing countries lack the resources to install or upgrade EWS infrastructure (radar, sensors, communication tools) and to train personnel. Fragmented financing is a problem. Support comes from various donors and programs without a unified strategy, leading to potential overlaps in some areas and stark gaps in others. For instance, recent analyses show that a large share of EWS funding is concentrated in a few countries, while Small Island Developing States (SIDS) and Least Developed Countries (LDCs) remain underfunded despite being highly vulnerable\footnote{See, \url{https://wmo.int/media/news/tracking-funding-life-saving-early-warning-systems}.}. Even when initial capital is provided to set up an EWS, securing long-term funding for operations and maintenance (software updates, staffing, equipment calibration) is difficult. Without sustainable financing, systems can degrade over time. Ensuring financial sustainability, co-financing arrangements, and political commitment is critical so that EWS are not one-off projects but enduring services.

In addition to the above, there are challenges in technological adoption and last-mile delivery: for example, reaching remote or marginalized populations with warnings (issues of language, literacy, and reliable communication channels) and building trust so that people heed warnings. Climate change is also introducing new complexities – hazards are becoming more unpredictable or intense, testing the limits of existing early warning capabilities. Overall, addressing data and standardization issues, improving institutional coordination, and closing funding gaps are priority challenges to fully realize the life-saving potential of EWS.

\subsection{Relevance to this study}

Our work is focused on the financial tracking and classification of investments in climate resilience, and EWS represent a prime example of such investments. Early warning projects often cut across sectors and funding sources – they might include components of infrastructure, technology, capacity building, and community outreach. Because of this cross-cutting nature, tracking where and how money is spent on EWS can be difficult without a clear classification system. Different organizations may label EWS-related activities in various ways (e.g. “hydromet modernization”, “disaster preparedness”, “climate services”), leading to inconsistencies in investment data. By establishing a standardized framework to define and categorize EWS investments, the study helps create a “big-picture view” of early warning financing. This enables analysts and policymakers to identify overlaps, gaps, and trends that were previously obscured by fragmented data.

Moreover, improving the classification of EWS funding directly supports broader resilience initiatives. For instance, the newly launched Global Observatory for Early Warning System Investments is already working to tag and track EWS-related expenditures across major financial institutions. Such efforts mirror the goals of this study by highlighting the need for consistent tracking, transparency, and coordination in climate resilience finance. Better classification of investments means stakeholders can pinpoint where resources are going and where additional support is needed to meet global targets like the “Early Warnings for All by 2027” pledge. In short, EWS feature in this study as a critical category of climate resilience investment that must be clearly identified and monitored.

By including EWS in its financial tracking framework, the study provides valuable insights for decision-makers. It helps determine how much funding is allocated to early warnings, from which sources, and for what components (equipment, training, maintenance, etc.). This information is crucial for evidence-based decisions on scaling up EWS: for example, spotting a shortfall in community-level preparedness funding, or recognizing successful investment patterns that could be replicated. Ultimately, linking EWS to the study’s financial tracking reinforces the message that climate resilience investments can be better managed when we know their size, scope, and impact area. By classifying EWS expenditures systematically, the study contributes to stronger accountability and strategic planning in building climate resilience, ensuring that early warning systems – and the communities they protect – get the support they urgently need.

\section{Dataset Construction}
In this study, we analyze financial information extracted from PDFs containing both structured and unstructured data. Unlike conventional benchmark datasets, these documents exhibit high heterogeneity in their formats—some tables are well-structured, while others embed financial figures within free-text paragraphs or are scattered across multiple rows and columns. Additionally, many numerical values correspond to multiple rows within the same column, creating challenges in extraction, alignment, and interpretation.

The annotated data, provided by experts in CSV format, along with the corresponding PDFs, can be found in the supplementary materials of this paper.

The dataset consists of 298 rows of expert annotations and contains the following 9 columns:  
\textit{Fund, Project ID, Component, Outcome/Expected-Outcome/Objectives, Output/Sub-component, Activity/Output Indicator, Page Number, Amount,} and \textit{Label}.  

The total amount of Early Warning Systems (EWS) is computed as the sum of all \textit{Amount} values for a given project.

The annotated dataset (CSV file and PDFs) consists of financial reports and investment documents sourced from publicly available institutional records, which are intended for public information and research and transparency purposes. The dataset is used strictly within its intended scope—analyzing financial tracking in climate investments—and adheres to the original access conditions. Additionally, for the artifacts we create, including benchmark datasets and classification models, we specify their intended use for research and evaluation in automated financial tracking and ensure they remain compliant with ethical research guidelines.

\end{document}